\documentclass[letterpaper]{article} 
\usepackage{aaai2026}  
\usepackage{times}  
\usepackage{helvet}  
\usepackage{courier}  
\usepackage[hyphens]{url}  
\usepackage{graphicx} 
\usepackage{natbib}  
\usepackage{caption} 
\usepackage{amsmath, amsthm, mathtools, amssymb, amsfonts}
\usepackage{nicefrac}
\usepackage{microtype}
\usepackage{xspace}
\usepackage{enumitem}
\usepackage{subcaption}
\usepackage{adjustbox}
\usepackage{multirow}
\usepackage{makecell}
\usepackage{siunitx}
\usepackage[capitalize,noabbrev]{cleveref}
\usepackage{overpic}
\usepackage{booktabs} 
\usepackage{algorithm}
\usepackage{algorithmic}
\usepackage{tikz}
\usepackage{xr}          

\newcommand{\ours}{FedMerge\xspace}

\title{FedMerge: Federated Model Merging for Personalization}

\author{
    Shutong Chen\textsuperscript{\rm 1},
    Tianyi Zhou\textsuperscript{\rm 2},
    Guodong Long\textsuperscript{\rm 1}\thanks{Corresponding author.},
    Jing Jiang\textsuperscript{\rm 1},
    Chengqi Zhang\textsuperscript{\rm 3}
}
\affiliations{
    \textsuperscript{\rm 1}University of Technology Sydney, Australia\\
    \textsuperscript{\rm 2}University of Maryland, College Park, USA\\
    \textsuperscript{\rm 3}The Hong Kong Polytechnic University, China\\
    shutong.chen@student.uts.edu.au, tianyi@umd.edu, guodong.long@uts.edu.au, jing.jiang@uts.edu.au, chengqi.zhang@polyu.edu.hk
}
\nocopyright
\begin{document}

\maketitle

\begin{abstract}
 One global model in federated learning (FL) might not be sufficient to serve many clients with non-IID tasks and distributions. Despite recent advances in FL to train multiple global models for better personalization, they only provide limited model choices to clients, so local finetuning of multiple models is still indispensable. This paper proposes a novel ``\ours'' approach that can create a single personalized model per client by simply merging multiple global models with automatically optimized and customized weights. We formulate this problem as a joint optimization of global models and the merging weights per client. Unlike existing FL approaches, where the server broadcasts one or multiple global models to all clients, the server only needs to send a customized, merged model to each client. Moreover, instead of periodically interrupting the local training and re-initializing it to a global model, the merged model aligns better with each client's task and data distribution, smoothing the local-global gap between consecutive rounds caused by client drift. We evaluate \ours on different non-IID settings applied to various domains with diverse tasks and data types, in which \ours consistently outperforms existing FL approaches, including clustering-based and mixture-of-experts (MoE) based methods. \looseness=-1
\end{abstract}

\begin{links}
    \link{Code}{https://github.com/shutong043/FedMerge}
\end{links}

\section{Introduction}

Federated learning (FL) enables decentralized and collaborative learning of models across various clients without sharing their local data. 
By exchanging only local models instead of raw data, FL ensures data privacy and security while leveraging distributed datasets for improved performance. In particular, local datasets across clients are often non-identically and independently distributed (non-IID), making the design of algorithms to address such non-IIDness a fundamental challenge.\looseness=-1

\begin{figure*}[ht]
    \centering
    \begin{subfigure}[t]{0.49\textwidth}
        \centering
        \includegraphics[width=\textwidth]{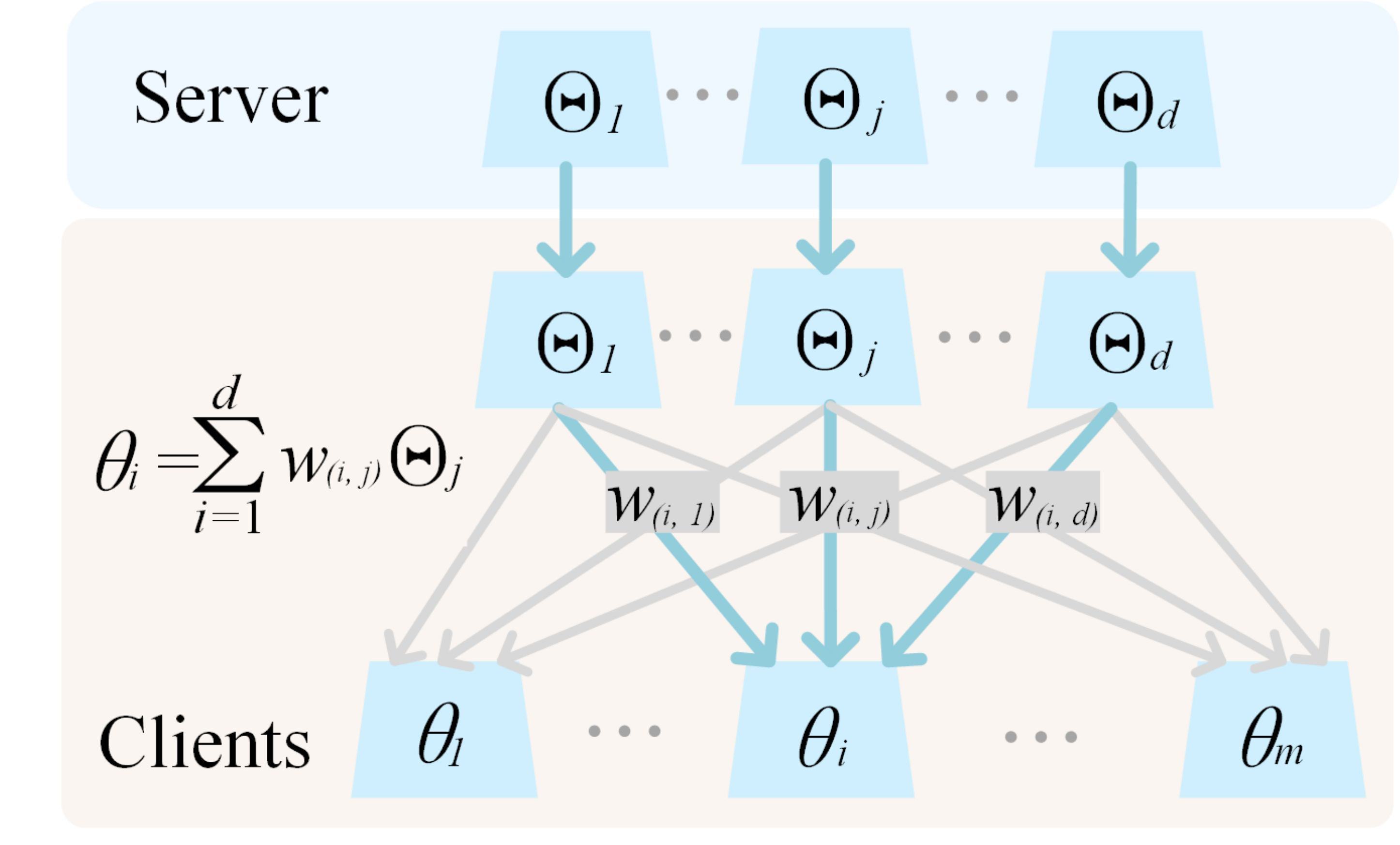}
        \caption{The architecture of \textbf{Federated Learning with Mixture of Experts}. Each client receives $d$ global models from the server and optimizes them separately. The client's resource consumption is proportional to the number of global models, making it inefficient.}
        \label{fig:fedmoe}
    \end{subfigure}
    \hfill
    \begin{subfigure}[t]{0.49\textwidth}
        \centering
        \includegraphics[width=\textwidth]{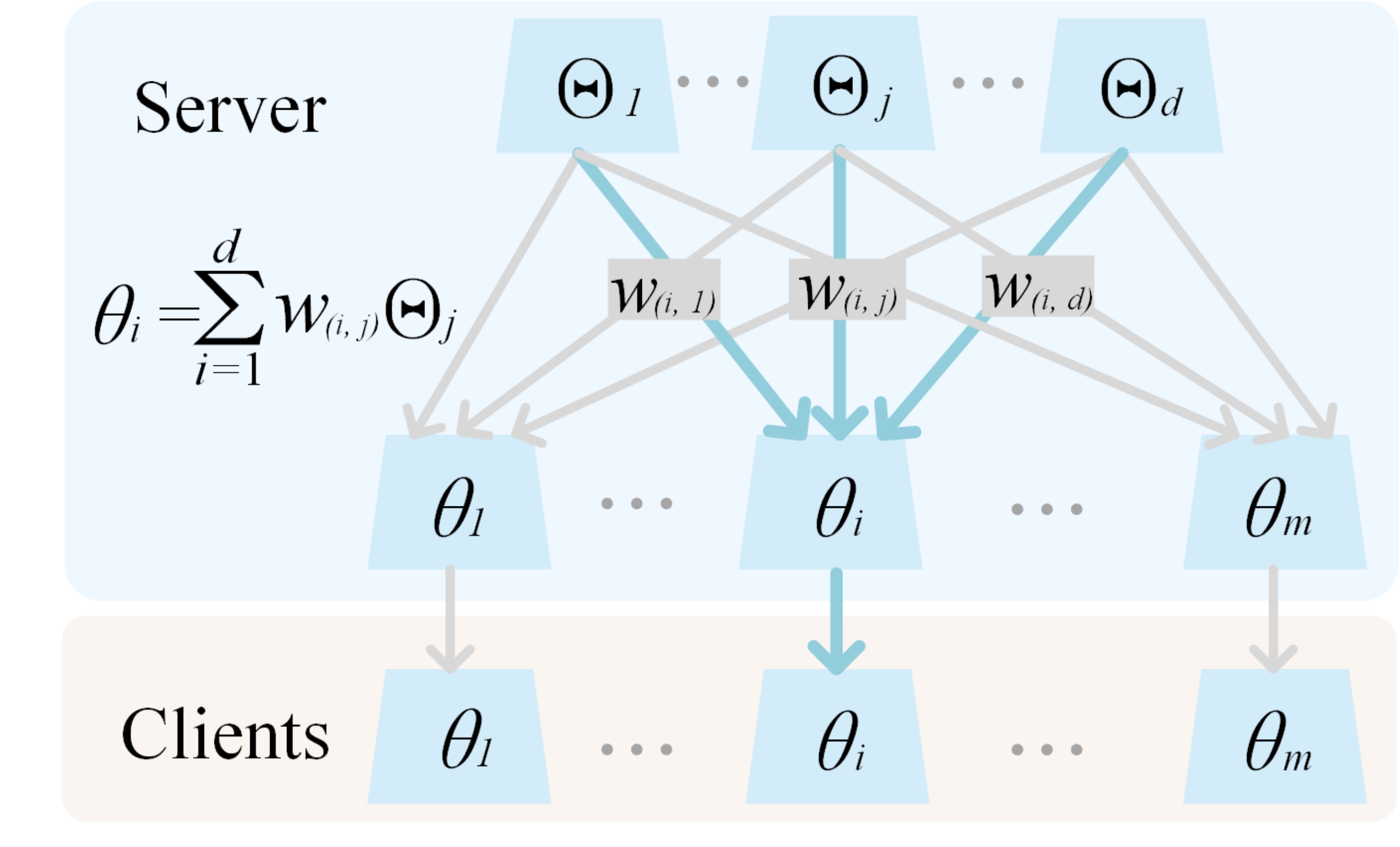}
        \caption{The architecture of the proposed \textbf{Federated Learning with Model Merging (\ours)}. The merging operation (weight averaging) is performed on the server side instead of on each client. Only the merged models, 
$\theta$, are communicated and optimized, making the cost irrelevant to the number of global models.}
        \label{fig:architecture}
    \end{subfigure}
    
    \caption{Comparison between MoE-like FL methods and the proposed \ours. Both methods perform weight averaging over global models. MoE-like FL performs weight averaging on each client, while \ours performs weight averaging on the server.}
    \label{fig:2x1}
\end{figure*}

Model merging \cite{yadav2024survey} is a knowledge ensemble technique that integrates multiple models into a single unified model. It has recently gained increasing attention in the context of foundation models \cite{lorahubhuang2023lorahub,chronopoulou2023adaptersoup,yang2024model,kim2025personalized,wang2025never}, as it enables efficient knowledge integration across diverse domains. Instead of training a large model from scratch, model merging combines existing pre-trained models into an enriched knowledge space, allowing effective reuse of previously trained experts. This flexibility naturally extends to parameter-efficient tuning methods, such as Low-Rank Adaptation (LoRA), which can be merged with minimal computational overhead. Due to these efficiency advantages, LoRA has been widely explored in FL settings \cite{yi2023fedlora,yang2024dualpersonal,sun2024improvingloraprivacy,cho2024heterogeneouslora}. These properties make model merging a promising approach for scaling and personalizing federated learning.\looseness=-1

Federated learning with multiple models is an important direction for addressing client heterogeneity \cite{fedemmarfoq2021federated,wu2023personalizedgaussianmixture}, and a recent dominant approach in this line of work is Mixture-of-Experts (MoE) \cite{luo2024mixture,feng2025pm,ijcai2025p610,radwan2025feddg}. In these MoE-like FL approaches, the server maintains a “model soup” of diverse experts that are shared across clients. As illustrated in Fig.~\ref{fig:fedmoe}, each client downloads multiple (or all) expert models and combines them locally to form a personalized model. However, sharing multiple models across clients introduces an intrinsic conflict: the desire for a rich, diverse set of global models conflicts with the limited computational and communication resources of individual clients. The cost for each client scales with the number of models it downloads, making it impractical to leverage a large model soup. While selecting a subset of models for each client can alleviate this issue \cite{chronopoulou2023adaptersoup}, it often requires task-specific knowledge, which limits its generalization ability. \looseness=-1

Model merging provides an elegant solution to the scalability issues faced by MoE-like FL methods. By integrating multiple expert models into a single personalized model, model merging allows each client to benefit from the richness and diversity of the model soup without having to download or train multiple models locally. This fundamentally eliminates the linear growth of communication and computation costs on the client side, making it feasible to leverage a large number of global models even in resource-constrained environments. \looseness=-1

Building on this insight, we propose \textbf{Federated Learning with Model Merging (\ours)}. As shown in Fig.~\ref{fig:architecture}, the server merges global models using client-specific weights to construct a unique personalized model for each client. Clients then only need to communicate and optimize a single merged model, while the server handles the merging operation with its typically greater resources. A challenge introduced by this design is how to update both the global models and the merging weights when only merged models are trained on clients. To address this, we formulate FedMerge as a joint optimization problem and derive a backpropagation-like update strategy to enable end-to-end training for both components. Experimental results on various non-IID settings show that \ours consistently outperforms clustering-based and MoE-based FL approaches.

We summarize the main contributions of this paper:
\begin{itemize}
    \item We propose a novel FL framework to solve the conflict between using multiple global models for personalization and the limited resources of clients.
    \item We introduce the Federated Merging (\ours) algorithm, where the server provides a personalized model to each client by merging a set of global models with client-specific weights.
    \item Convergence analysis is provided in the Appendix~\ref{sec:convergence_analysis}.
    \item Experimental results demonstrate that \ours offers a cost-effective way to improve performance by supporting multiple global models while maintaining a constant and low client-side cost.
\end{itemize}
\section{Related Work}

\textbf{Model Merging}
Model merging refers to methods aimed at reusing the knowledge contained in multiple models. The fundamental technique of model merging is to integrate existing models to form new ones. There have been a number of works focusing on merging methodologies, such as simple averaging \cite{modelsoupwortsman2022model}, merging with Fisher Information \cite{fishermergingmatena2022merging}, and merging with task vectors \cite{taskvectorilharco2022editing}. Recently, model merging has become closely related to the development of foundation models \cite{lu2023routing,lu2024twin,ostapenko2024towards,zhou2025mergeme,cheng2025whoever,du2025adamms}, as such developments have brought numerous reusable expert foundation models online \cite{huggingface}. For example, LoraHub \cite{lorahubhuang2023lorahub} trains one LoRA expert per task on a collection of 200 tasks and combines these experts to evaluate downstream tasks. Due to the large number of expert models, model selection plays a practical role in choosing the most relevant experts for each downstream task \cite{chronopoulou2023adaptersoup,zhao2024retrieval}. Model merging is also related to Mixture of Experts (MoE)-based methods, where the design of routing weights is a key factor. The router can be neural network-based \cite{lu2023routing}, feature similarity-based \cite{loraflowwang2024lora}, or even SVD-decomposed \cite{ostapenko2024towards}. In this paper, rather than directly adopting the various model merging techniques mentioned above in FL, we leverage the fundamental concept of model merging to address the primary challenge in Federated Learning with Multiple Models, enabling a large model soup to be efficiently reused on the server while communicating only a single model to each client. \looseness=-1

\textbf{Federated Learning with Multiple Models}
The use of multiple models in FL arises from the fact that sharing a single model cannot effectively address the Non-IID problem. Personalized FL methods \cite{t2020personalized,li2021ditto,collins2021exploiting} apply the concept of multiple models by designing a unique model for each client's personalized data distribution. Our method is more related to a type of multi-model FL, where multiple models are shared on the server \cite{fedemmarfoq2021federated,wu2023personalizedgaussianmixture}. A recent line of research explores mixture-of-experts (MoE) architectures in FL \cite{luo2024mixture, feng2025pm,ijcai2025p610,radwan2025feddg,xie2025dflmoe}, where multiple experts and personalized gating mechanisms are introduced to enhance personalization and handle strong data heterogeneity. These approaches either deploy MoE models on the server to improve scalability or distribute experts across clients for decentralized adaptation, enabling dynamic expert selection and robust personalization across domains. For example, pFedMoE \cite{yi2024fedmoe} employs one local expert and one global expert for each client, while deriving routing weights using personalized router neural networks. Recently, a few methods have incorporated merging concepts into FL \cite{chen2024local,chen2025breaking}. In \cite{tao2024task, liu2024fedlpa}, a one-shot federated learning scenario is considered, and model merging methods can be naturally extended to address the challenges rise in this setting.
\cite{salami2024closed} proposed an end-to-end optimization framework that leverages RegMean \cite{jin2022dataless} to constrain the update distribution of LoRA adapters. While these approaches also utilize model merging to improve federated learning, our work targets a different challenge within this paradigm: balancing the scalability of a large pool of global models with the limited communication and computation capacities of individual clients. \looseness=-1

\section{Backgrounds and Motivation}

\subsection{Preliminaries}

Federated Learning (FL) enables multiple clients to collaboratively train a model under the coordination of a central server without sharing their private data. Let $(X_i, Y_i)$ denote the data of client $i$. In standard FL, the goal is to train a single global model $\Theta$ that minimizes the weighted sum of local objectives across all $m$ clients:
\begin{equation}\label{eq:FL}
\begin{aligned}
   & \min_{\Theta}\sum_{i=1}^m\frac{n_i}{n}\ell\left(Y_i,f(X_i; \Theta)\right)
\end{aligned}
\end{equation}
where $n_i$ and $n$ denote the data sizes of client $i$ and all clients, respectively. This formulation, known as FedAvg \cite{mcmahan2017communication}, aggregates locally trained models to form a unified global model. However, under non-IID data distributions, a single global model often fails to generalize well to each client’s local domain.

In parallel, the concept of a model soup \cite{modelsoupwortsman2022model} refers to a collection of trained models whose parameters are combined—typically via simple averaging or linear interpolation—to improve generalization performance. The core idea is that interpolating between diverse models trained from different optimization trajectories can reveal complementary information, thereby enhancing both robustness and overall generalization.

\subsection{Problem Setting}

To address the limitations of FedAvg under heterogeneous data, personalized federated learning (PFL) allows each client to maintain a customized model $\theta_i$, optimized with respect to its local data: 
\begin{equation}\label{eq:PFL}
\begin{aligned}
   & \min_{\theta}\sum_{i=1}^m\frac{n_i}{n}\ell\left(Y_i,f(X_i; \theta_i)\right)
\end{aligned}
\end{equation}
where each $\theta_i$ learns client-specific knowledge while still being related to some shared global representation.

Motivated by the insight from model soup, we consider how this notion can be adapted to the PFL setting, where the server can maintain a shared model soup composed of models aggregated or collected from different clients. This shared model soup enables model construction that balance personalization and generalization.

\section{Methodology}

\subsection{Objective Function}
We place a model soup containing multiple models on the server. Let $\{\Theta_1, \Theta_2, \ldots, \Theta_d\}$ denote a set of $d$ global models, where each model is trained with respect to its own knowledge space. In federated learning (FL), each client is associated with a personalized task, and we assume that the knowledge required to solve each task lies within the set of global models maintained on the server, that is, within the shared model soup.  \looseness=-1

Therefore, the server can create a personalized model for each client by reusing the global models in the model soup. In this work, we adopt standard weight merging, where the global models are combined with task-specific merging weights to generate a personalized model for each client:
\[
\theta_i = \sum_{j=1}^d w_{(i,j)} \cdot \Theta_j,
\]
where $w_{i,1:d}$ represents the personalized merging weights for client $i$, determining the contribution of each global model to the merged model.

After this merging operation on the server, the personalized merged models are sent to the clients. The objective of FedMerge is formulated as:
\begin{equation}\label{eq:MGFL_obj}
\min_{\theta} \sum_{i=1}^m \frac{n_i}{n} \, \ell \left(Y_i, f(X_i; \theta_i)\right)
\end{equation}
\begin{equation}\label{eq:MGFL_constraint}
\text{s.t.} \quad \theta_i = \sum_{j=1}^d w_{(i,j)} \cdot \Theta_j.
\end{equation}

Note that in Eq~\eqref{eq:MGFL_constraint}, each client is optimized only on its merged model $\theta_i$, rather than directly on the global models or the merging weights. This is a key difference between FedMerge and MoE-like FL methods.

\subsection{Optimizing via Backpropagation-like Strategy}

Solving Eq~\eqref{eq:MGFL_obj} with Eq~\eqref{eq:MGFL_constraint} requires optimizing $w$ and $\Theta$. In the proposed \ours, each client only receives the merged model, and only the gradient of the merged model $\frac{\partial \ell}{\partial \theta}$ is directly available. The weights $w$ and global models $\Theta$ remain on the server, and their gradients cannot be computed directly.\looseness=-1

Inspired by the mechanism of backpropagation in neural networks \cite{rumelhart1986learning}, we propose an update rule for merging weights $w$ and global models $\Theta$.
If we consider FedMerge as a single-layer fully connected network, where local models $\theta$ are high-level nodes and global models $\Theta$ are low-level nodes, then $w$ acts as the propagation weights. The gradient of $w$ and $\Theta$ depends on the gradient of $\theta$. Following the chain rule, we compute $\frac{\partial L}{\partial w}$ and $\frac{\partial L}{\partial \Theta}$ using $\frac{\partial L}{\partial \theta}$ as an intermediate variable: 

\begin{equation}
\frac{\partial L}{\partial \Theta_j} = \sum_{i=1}^m \frac{n_i}{n}w_{(i,j)}\frac{\partial \ell}{\partial \theta_i}.
\label{eq:gradient_theta}
\end{equation}
\begin{equation}
\frac{\partial L}{\partial w_{(i,j)}} = \frac{n_i}{n} \langle \Theta_j, \frac{\partial \ell}{\partial \theta_i} \rangle.
\label{eq:gradient_w}
\end{equation}

Eq~\eqref{eq:gradient_theta} and Eq~\eqref{eq:gradient_w} are derived directly from the objective of \ours in Eq~\eqref{eq:MGFL_obj} with Eq~\eqref{eq:MGFL_constraint}. The key factor is the use of local gradient $\frac{\partial L}{\partial \theta}$ as an intermediate variable. For the detailed derivation process, please refer to Appendix~\ref{sec:gradient_derivations}. 

\begin{figure}[t]
    \centering
    \begin{overpic}[width=0.95\linewidth]{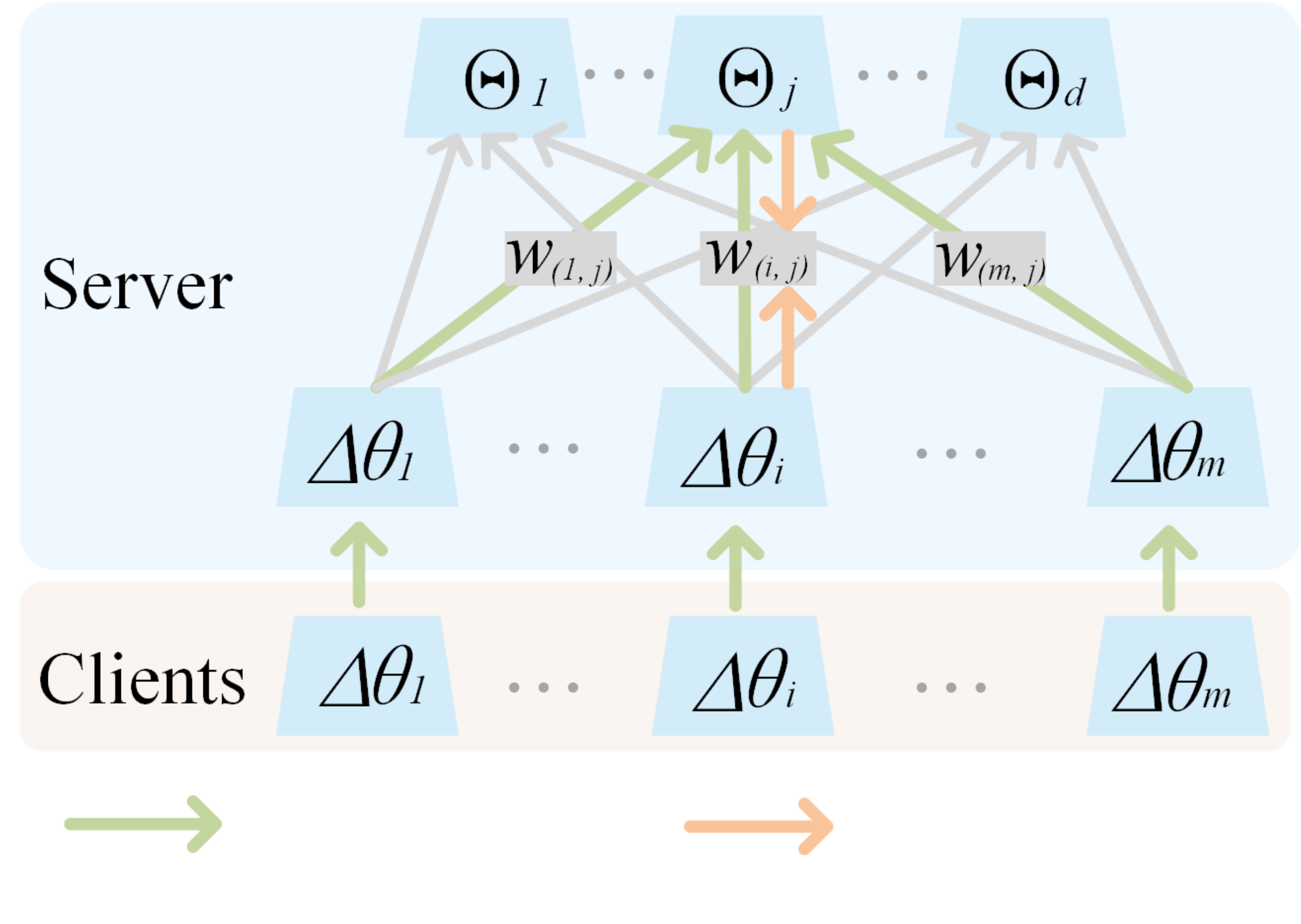}
        \put(20, 8){\small Global models}
        \put(20, 3){\small update: Eq~\eqref{eq:update_Theta_delta}, Eq~\eqref{eq:update_Theta}}
        \put(66, 8){\small Merging weights}
        \put(66, 3){\small update: Eq~\eqref{eq:update_w_delta}, Eq~\eqref{eq:update_w}}
    \end{overpic}
    \caption{The update information flow of global models and merging weights in FedMerge. Global models are updated from all clients, and merging weights are updated from their associated model pairs.}
    \label{fig:FedMerge_update}
\end{figure}

To better illustrate Eq~\eqref{eq:gradient_theta} and Eq~\eqref{eq:gradient_w}, Fig.~\ref{fig:FedMerge_update} highlights the information flows from local models to global models and merging weights. It shows that the gradient of each global model is influenced by all client updates, while the updates for each merging weight depend on both the global model and the merged model it connects.

We summarize the training procedure as follows:

\begin{enumerate}[wide, label=\textbf{\arabic*.}]

    \item \textbf{Forward Propagation:}  
    The server computes the merged models for each client using Eq~\eqref{eq:MGFL_constraint} and distributes them accordingly. Each client then updates its merged model based on its local objective function:
    
    \begin{equation}\label{eq:local_update}
        \theta^{(t)}_{i} - \theta^{(t-1)}_{i} \leftarrow - \eta \nabla_{\theta^{(t-1)}_{i}} \ell(Y_i, f(X_i; \theta^{(t-1)}_{i})),
    \end{equation}

    where $\eta$ is the learning rate for $\theta$. Notably, the local update procedure in FedMerge is the same as in FedAvg. 

    \begin{algorithm}
    \caption{Federated Merging (\ours)}
    \label{alg}
    \begin{algorithmic}[1]
    \STATE \textbf{Initialize:} global models $\{\Theta_1,\Theta_2,\ldots,\Theta_d\}$, merging weights $w \in \mathbb{R}^{m \times d}$
    \FOR{each communication round}
        \STATE Randomly select a subset $A$ of clients
        \FOR{each client $i \in A$}
            \STATE Merge models using Eq~\eqref{eq:MGFL_constraint} to obtain $\theta_i$
            \STATE Send merged model $\theta_i$ to client $i$
        \ENDFOR
        \FORALL{selected clients (in parallel)}
            \STATE Update $\theta_i$ using local gradient descent via Eq~\eqref{eq:local_update} for $t$ steps
            \STATE Send local update $\Delta\theta_i$ to the server
        \ENDFOR
            \STATE Compute gradients for $\Theta$ via Eq~\eqref{eq:update_Theta_delta}
            \STATE Update $\Theta$ via Eq~\eqref{eq:update_Theta}
            \STATE Compute gradients for $w$ via Eq~\eqref{eq:update_w_delta}
            \STATE Update $w$ via Eq~\eqref{eq:update_w}
    \ENDFOR
    \end{algorithmic}
    \end{algorithm}
    
    \item \textbf{Backward Propagation:}  
    After the local update, each client transmits its 
    updated parameters $\Delta\theta_i = \theta^{(t)}_{i} - \theta^{(0)}_{i}$ back to the server. The server then updates both the global models $\Theta$ and the merging weights $w$. Using the chain rule in Eq~\eqref{eq:gradient_theta} and Eq~\eqref{eq:gradient_w}, the updates for the merging weights and global models are computed based on local updates:
    
    \begin{equation}
    \Delta \Theta_j = \sum_{i=1}^m \frac{n_i}{n} w_{(i,j)} \Delta \theta_i,
    \label{eq:update_Theta_delta}
    \end{equation}
    
    \begin{equation}
    \Delta w_{(i,j)} = \frac{n_i}{n} \langle \Theta_j, \Delta \theta_i \rangle.
    \label{eq:update_w_delta}
    \end{equation}
    \item \textbf{Parameter Update:}  
    Once the gradients for the global models and merging weights are obtained, the server updates $\Theta$ and $w$:
    
    \begin{equation}
    \Theta_j \leftarrow \Theta_j + \Delta \Theta_j,
    \label{eq:update_Theta}
    \end{equation}
    
    \begin{equation}
    w_{(i,j)} \leftarrow w_{(i,j)} + \Delta w_{(i,j)}.
    \label{eq:update_w}
    \end{equation}

    The overall \ours process is summarized in Algorithm~\ref{alg}. We also provide detailed convergence analysis, please refer to Appendix~\ref{sec:convergence_analysis} for details.
\end{enumerate}

\begin{table*}[t]
\centering
\renewcommand{\arraystretch}{0.95} 
\footnotesize
\resizebox{0.9\textwidth}{!}{%
\begin{tabular}{llcccccc}
\toprule
\multicolumn{8}{l}{\textbf{Classic FL (ResNet-based)}} \\
\midrule
\multirow{2}{*}{\textbf{Multi-Model}}     
& \textbf{Model Num.}          & ResNet-9×5 & ResNet-9×10 & ResNet-9×15 & ResNet-9×20 & ResNet-9×25 & ResNet-9×30 \\ 
& \textbf{Parameter Scale}     & 5 & 10 & 15 & 20 & 25 & 30 \\
\multirow{2}{*}{\textbf{Single-Model}}   
& \textbf{Model Arch.}         & ResNet-9 & ResNet-18 & ResNet-34 & ResNet-50 & ResNet-101 & ResNet-152 \\ 
& \textbf{Parameter Scale}     & 1 & 2.27 & 4.32 & 4.82 & 8.73 & 11.95 \\
\midrule
\multicolumn{8}{l}{\textbf{Foundation Model FL (LoRA-based)}} \\
\midrule
\multirow{2}{*}{\textbf{Multi-Model}} 
& \textbf{Description}         & Rank-8×2 & Rank-8×4 & Rank-8×6 & Rank-8×8 & Rank-8×10 & Rank-8×12 \\
& \textbf{Parameter Scale}     & 2        & 4        & 6        & 8        & 10        & 12        \\
\multirow{2}{*}{\textbf{Single-Model}} 
& \textbf{Description}         & Rank-8   & Rank-16  & Rank-24  & Rank-32  &           &           \\
& \textbf{Parameter Scale}     & 1        & 2        & 3        & 4        &           &           \\
\bottomrule
\end{tabular}%
}
\caption{Parameter scale settings for classic FL (top) and foundation model FL (bottom). The scale is measured relative to Rank-8 for LoRA modules and to ResNet-9 for ResNet architectures.}
\label{tab:merged_param_scale}
\end{table*}

\subsection{Technical Implementations}

\textbf{Constraint on Merging Weights Values} 
The original objective of \ours in Eq~\eqref{eq:MGFL_constraint} has no constraints on merging weights, allowing the merging weight values to be negative. However, neural network models typically require regularized parameters. To address this, we apply a softmax function to the merging weights to obtain normalized weights. This introduces slight modifications to the updates in Eq~\eqref{eq:update_Theta_delta} and Eq~\eqref{eq:update_w_delta}. Detailed derivations can be found in Appendix~\ref{sec:softmax_gradient_derivations}.

\textbf{Clarification of Merging Weights Gradient}
In Eq~\eqref{eq:update_w_delta}, the update of merging weights involves the inner product of model parameters and gradients. For large models, the effective update directions can be overwhelmed by the inner products of a large number of parameters. To mitigate this issue, we simply use the parameters from the last few layers of the model for inner product—specifically, the classification head in convolutional neural networks. This is because higher-level layers capture more semantic information \cite{yosinski2014transferable}, and their parameters or gradients are correlated with task heterogeneity across clients. \looseness=-1

\subsection{Server-side Computational Complexity}

\textbf{FedMerge is as computational efficient on the server side as standard baselines} Server-side model assembly may raise concerns about computational overhead. We compare the server-side complexity of FedMerge with widely adopted baselines. As in Eq~\eqref{eq:MGFL_constraint}, Eq~\eqref{eq:update_Theta_delta} and Eq~\eqref{eq:update_w_delta}, FedMerge performs weight averaging \( d \) times over \( m \) models or computes \( md \) inner products, resulting in a complexity of \( \mathcal{O}(md) \). Cluster FL~\cite{ghosh2020efficientifca,long2023multi} and Multi-model FL~\cite{fedemmarfoq2021federated,bhuyan2022multi}—both widely accepted in the literature—also require performing weight averaging \( d \) times over \( m \) models on the server side, resulting in a complexity of \( \mathcal{O}(md) \). Thus, FedMerge does not introduce additional computational complexity on the server side compared to these standard approaches.

\section{Experiments}
\subsection{Classical FL Settings}

\textbf{Baselines}  
\ours relates to Cluster FL, Multi-model FL, and MoE-like FL. We compare it with the following representative methods:  
1) \textbf{Single-model} methods share a single global model, including Local (training local models separately on each client), FedAvg \cite{mcmahan2017communication}, FedAvg+ (FedAvg with local fine-tuning), and pFedMoE \cite{yi2024fedmoe}. Although named “MoE,” pFedMoE uses only one global model.   
2) \textbf{Multi-model} methods maintain multiple global models, such as IFCA \cite{ghosh2020efficientifca} (Cluster FL) and FedEM \cite{fedemmarfoq2021federated} (MoE-like FL).

\textbf{Parameter Scale}  
To ensure fair comparison, we control the parameter scale for each method. Multi-model methods (IFCA, FedEM, FedMerge) use ResNet-9 with varying numbers of global models from \{5, 10, 15, 20, 25, 30\}. Single-model methods (Local, FedAvg, FedAvg+, pFedMoE) adopt different ResNet variants: ResNet-9, ResNet-18, ResNet-34, ResNet-50, ResNet-101, and ResNet-152, whose parameter scales relative to ResNet-9 are \{1, 2.27, 4.32, 4.82, 8.73, 11.95\}. Full settings are shown in Table~\ref{tab:merged_param_scale}.

\textbf{\ours Settings}  
Global models are randomly initialized and trained from scratch. Merging weights are initialized uniformly to reflect equal initial contributions from global models.\looseness=-1

\textbf{System Settings}  
These are the universal settings for all methods across all non-IID settings. Each dataset is split into training, validation, and test sets, with the best checkpoint selected based on validation error. We use a constant learning rate of $0.01$, local training for $2$ epochs, and $500$ communication rounds. Each method is executed three times, and the reported results are the average of these runs.

\subsection{Parameter Usage vs. Performance}

\begin{figure}[t]
    \centering
    \begin{subfigure}{0.95\linewidth}
        \centering
        \includegraphics[width=\linewidth]{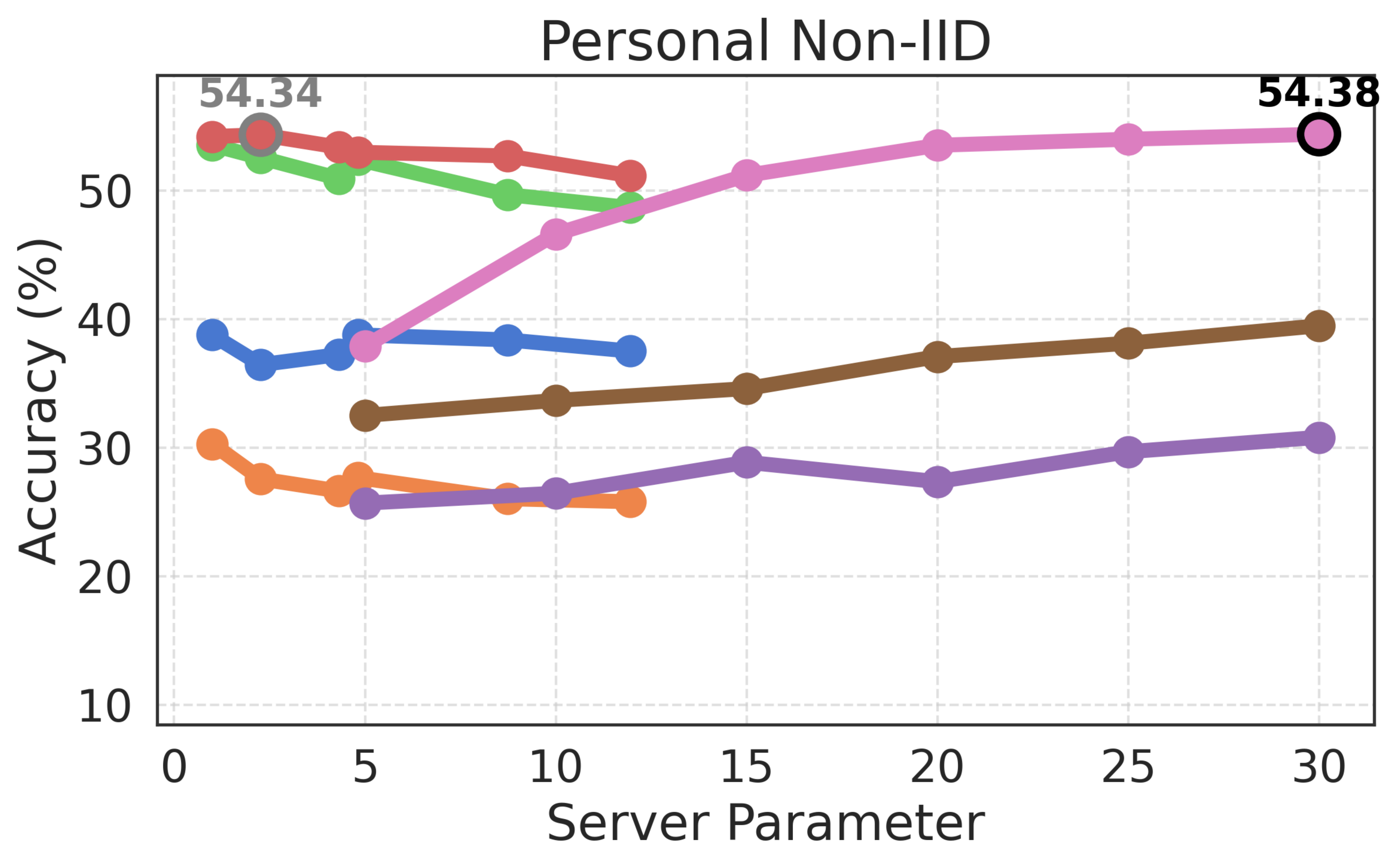}
        \caption{}
        \label{fig:server_personal}
    \end{subfigure}
    \begin{subfigure}{0.95\linewidth}
        \centering
        \includegraphics[width=\linewidth]{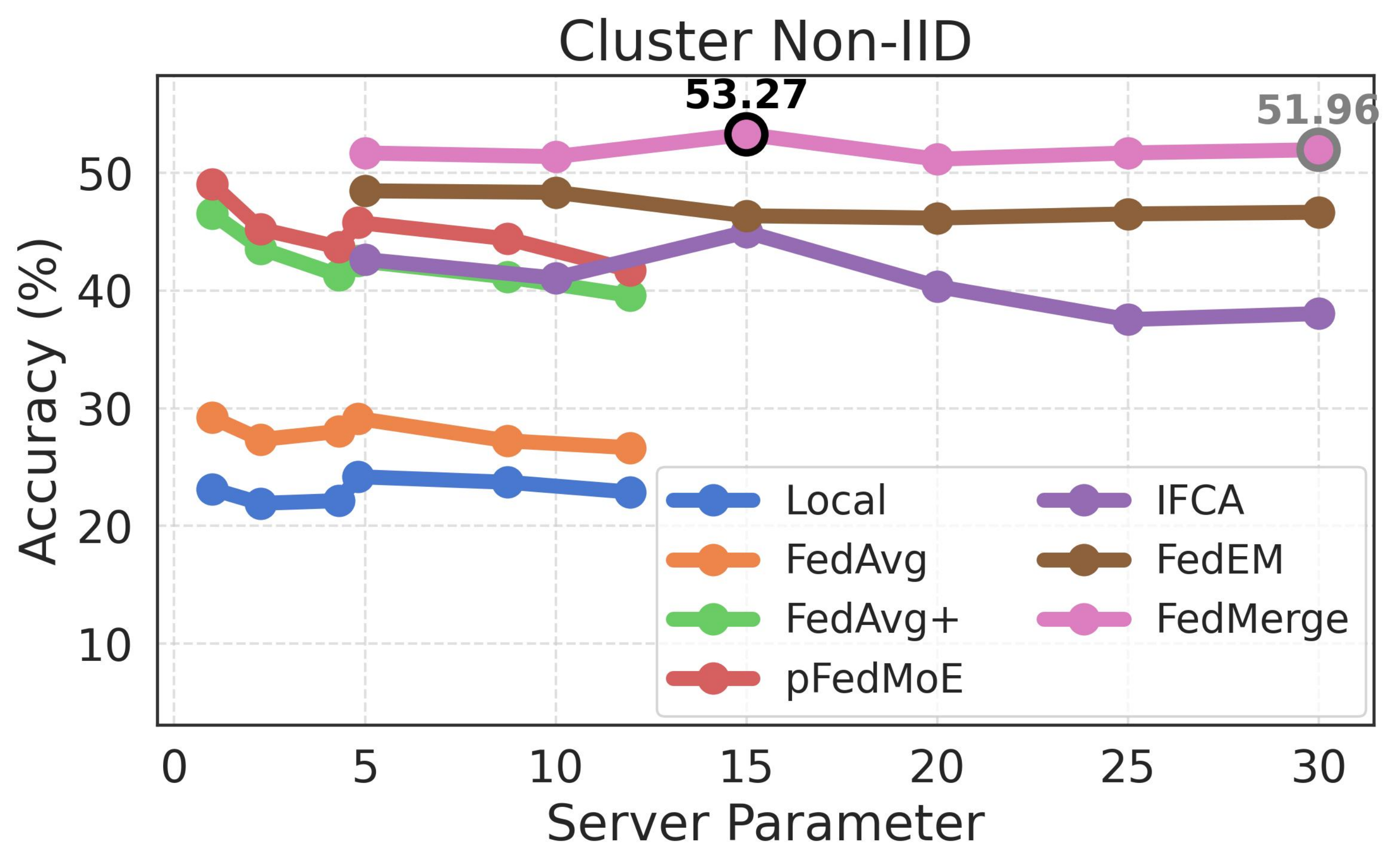}
        \caption{}
        \label{fig:server_cluster}
    \end{subfigure}

    \caption{Server parameter usage vs. accuracy. The x-axis represents multiples of ResNet-9's parameter count. For each method, the variation of server parameters follows Table~\ref{tab:merged_param_scale}. The highest and second-highest accuracy points are marked with black and gray respectively.}
    \label{personal_curve_share}
\end{figure}

\begin{figure*}[ht]
    \centering
    \begin{subfigure}[b]{0.49\textwidth} 
        \centering
        \includegraphics[width=\linewidth]{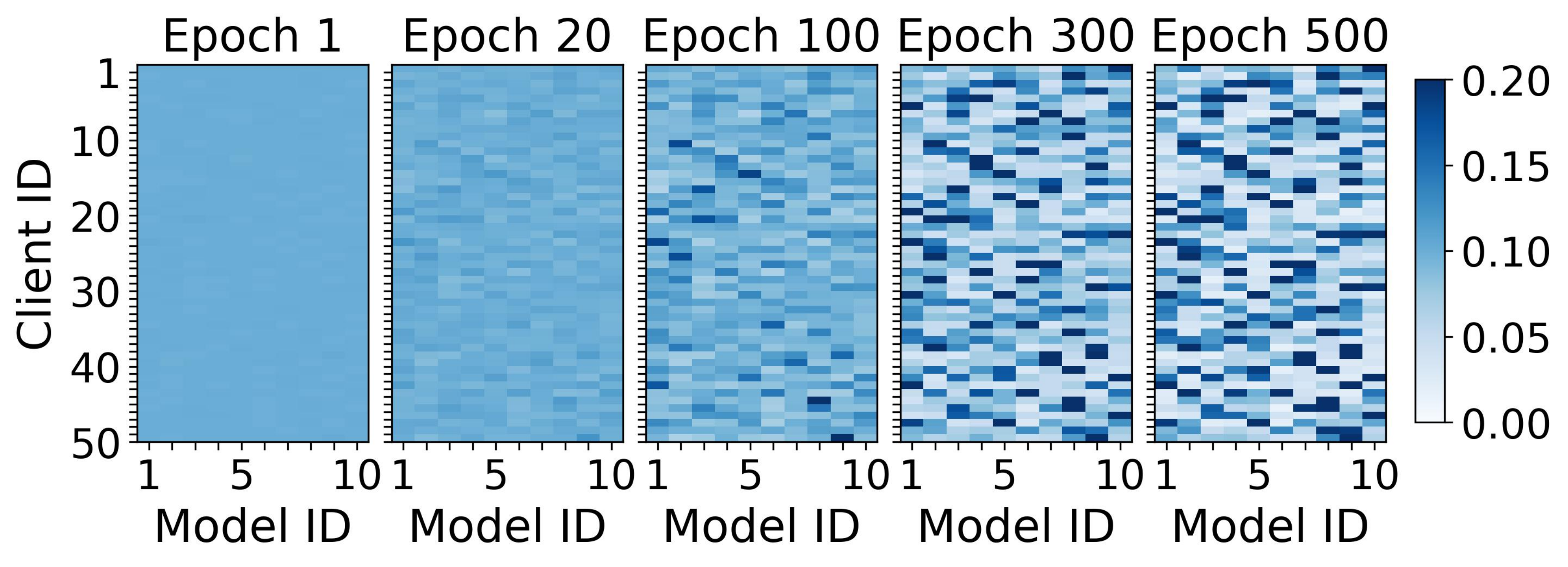} 
        \caption{Heat map of merge weights for Personal Non-IID.}
        \label{fig:sub1}
    \end{subfigure}
    \hfill 
    \begin{subfigure}[b]{0.49\textwidth} 
        \centering
        \includegraphics[width=\linewidth]{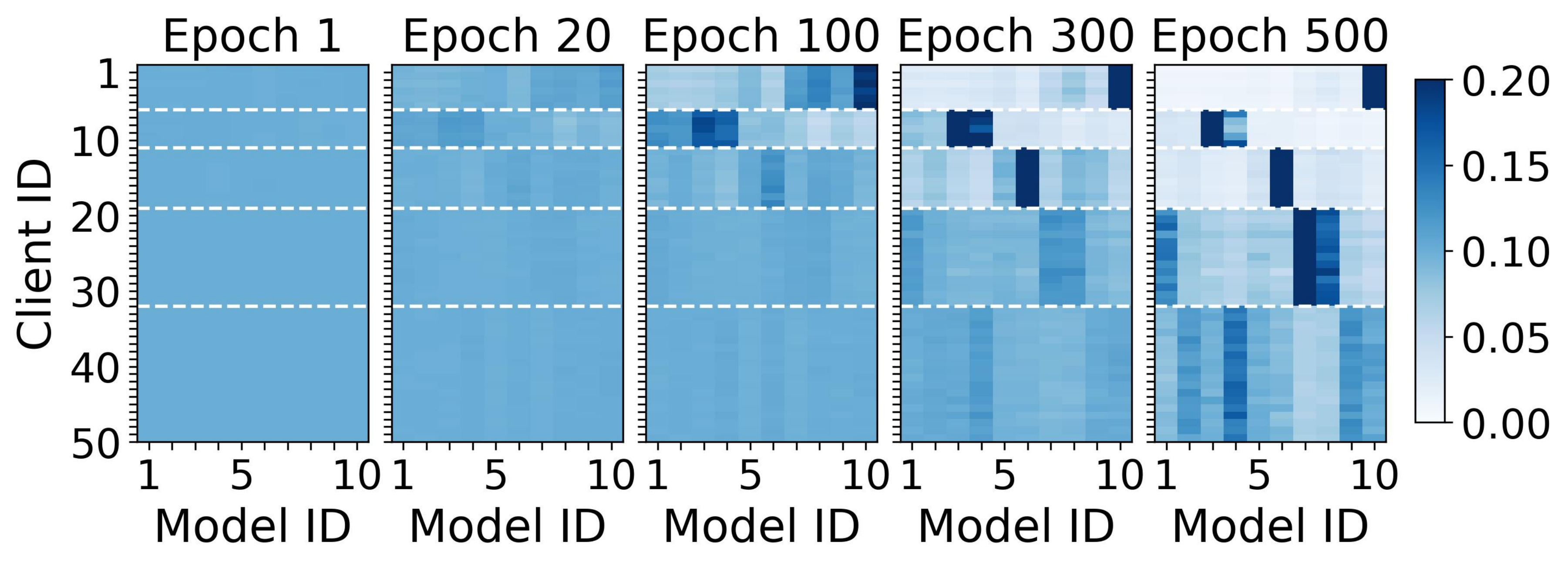} 
        \caption{Heat map of merge weights for Cluster Non-IID.}
        \label{fig:W strcutre cluster}
    \end{subfigure}

    \caption{Visualization of merge weights with 10 global models and 50 clients. Each row denotes one normalized weight vector for each client. In Cluster Non-IID, we separate ground-truth clusters with white dashed lines.}
    \label{fig:W strcutre}
    
\end{figure*}

\ours adapts to different Non-IID settings by adjusting the merging weights adaptively. To validate this adaptability, we consider two data distributions of different non-IID structures:

\textbf{Non-IID Settings}  
We build two non-IID settings using the CIFAR-100 \cite{krizhevsky2009learningcifar100} dataset:  \textbf{Cluster Non-IID}: We split the original CIFAR-100 dataset into $5$ clusters based on label categories. Each cluster is evenly assigned to a subgroup of clients out of 50 clients. The ground-truth clustering is \{\{1, …, 6\}, \{7, …, 11\}, \{12, …, 19\}, \{20, …, 32\}\}, \{33, …, 50\}\}.  
\textbf{Personal Non-IID}: A commonly used Non-IID setting proposed by \cite{hsu2019measuringnon-iid}, where the CIFAR-100 dataset is partitioned using a Dirichlet distribution into 50 clients. We set the concentration parameter $\alpha = 0.1$ for high Non-IIDness.

Following Table~\ref{tab:merged_param_scale}, we vary Single-model methods by model architecture and vary Multi-model methods by number of models, which results in various server parameter usage. As shown in Fig.~\ref{personal_curve_share}, we plot the server parameter usage vs. accuracy on Personal Non-IID and Cluster Non-IID settings. The results suggest the following:

\textbf{Multi-model FL methods provide an alternative way to improve performance beyond scaling model architectures.} In Fig.~\ref{fig:server_personal}, using larger ResNet architectures for Single-model methods leads to a decrease in performance, while using more ResNet-9 models for Multi-model methods can boost the performance (with the largest architecture considered being ResNet-512 for Single-model methods, and up to 30 models used for Multi-model methods). This is because both Personal and Cluster Non-IID settings involve 50 clients with sparse client data and high non-IID data partition settings, which limits performance gains from scaling up the model. In contrast, Multi-model methods can enhance the base model's ability to handle complex non-IID distributions by increasing the number of models.  \ours consistently surpasses FedEM and IFCA across various global model numbers, demonstrating the effectiveness of \ours compared to other Multi-Model FL methods. 

\textbf{\ours boosts performance by shifting computation to the server, leveraging the server’s abundant resources to accommodate the limited client-side capacity.} In Fig.~\ref{personal_curve_share}, FedMerge manages a number of ResNet-9 model on the server while transmitting and optimizing a single model per client. Notably, on Personal Non-IID setting, \ours outperforms pFedMoE under the Personal Non-IID setting $(54.38>54.34)$ with only one ResNet-9 per client, while pFedMoE needs two ResNet-18 models and additional router weights per client to achieve optimal performance. On Cluster Non-IID, FedMerge consistently surpasses other baselines and achieves optimal performance with $15$ global models.

We also provide Non-IID settings with PACS \cite{li2017deeper} dataset, where using larger ResNet architectures can boost performance. Additional results are in Appendix~\ref{sec:pacs}. \looseness=-1

\subsection{Ablation Study}

\textbf{\ours adapts to Non-IID settings with dynamic merging weights:}  
To understand the behavior of \ours under different Non-IID settings, we visualized the heatmaps of merging weight matrices in Fig.~\ref{fig:W strcutre}. Each row represents the normalized merging weights for a single client. Initially, the merging weight matrix is uniformly initialized, and as the epoch progresses, it begins to exhibit specific distribution patterns. 
We separate the ground-truth clusters in Cluster Non-IID with white dashed lines. In Cluster Non-IID, rows of merging weights are similar within each ground-truth cluster and distinct between different clusters. This indicates that \ours automatically allows clients with similar data distributions to prefer similar subgroups of global models, achieving a soft clustering. In contrast, for Personal Non-IID, rows of merging weights are relatively distinct from one another, and the merging weight matrix does not exhibit a regular structure.

\begin{table}[t]
\centering
\resizebox{0.95\linewidth}{!}{%
\begin{tabular}{lccccc}
\toprule
\multirow{2}{*}{Non-IID} & \multirow{2}{*}{FedAvg} & \multicolumn{4}{c}{\ours (15 global models)} \\ 
\cmidrule(lr){3-6}
                          &                         & Fixed (1/15) & Fixed (5/15) & Fixed (10/15) & Dynamic  \\ 
\midrule
Personal                  & 29.20          & 40.94 & 44.70 & 45.48 & \textbf{51.18} \\ 
Cluster                   & 30.29          & 32.55 & 32.52 & 34.06 & \textbf{53.27} \\ 
\bottomrule
\end{tabular}%
}
\caption{Ablation study on merging weights. Fixed indicates that each client randomly selects some global models, and the corresponding merging weight is uniformly initialized and frozen during training.}
\label{tab:merging_weight}
\end{table}

\begin{table*}[t]
\centering
\fontsize{9pt}{10pt}\selectfont
\resizebox{\textwidth}{!}{%
\begin{tabular}{lcccccccc|c}
\toprule
\textbf{Method} & \makecell[c]{Paraphrase} & \makecell[c]{Entailment} & \makecell[c]{Structure\\to text} & \makecell[c]{Text\\formatting} & \makecell[c]{Linguistic\\acceptability} & \makecell[c]{Word\\disambiguation} & \makecell[c]{Coreference} & \makecell[c]{Question\\classification} & \textbf{Average} \\
\midrule
FedEM    & 73.40 & 74.30 & \textbf{69.32} & \textbf{87.93} & 78.80 & \textbf{61.31} & 66.82 & 78.33 & 73.77 \\
FedMerge & \textbf{75.49} & \textbf{77.02} & 68.54 & 86.97 & \textbf{80.64} & 59.70 & \textbf{70.61} & \textbf{80.03} & \textbf{74.86} \\
\bottomrule
\end{tabular}%
}
\caption{ROUGE-1 scores (\%) of MoE-like method (FedEM) and model merging (FedMerge) in task-heterogeneous instruction following. Results are reported across 8 instruction-following tasks from Flan, with the average shown in the last column.}
\label{tab:fedem_fedmerge_taskwise}
\end{table*}

\textbf{Effect of different weight merging strategies:}  
We further analyze different weight merging strategies in Table~\ref{tab:merging_weight}. As observed, The performance of Fixed strategy decreases compared with dynamic merging, and Cluster FL decreases more than Personal Non-IID because random selection conflicts with the structural information of the cluster.

\subsection{Federated Finetuning of Large Language Models}
\textbf{Baselines} We use the same baselines as in classic FL settings, except that FedAvg is replaced with FedIT \cite{zhang2024towardsllmfedavg} and FedAvg+ is replaced with FedLoRA \cite{yi2023fedlora}. For pFedMoE and FedEM, we build personalized routers for the LoRA modules. Details can be found in Appendix~\ref{baseline_llm}.

\textbf{Datasets} Several baselines \cite{ye2024openfedllm,kuang2024federatedscope,ye2024fedllm} exist for federated foundation models. However, these models are primarily designed for instruction fine-tuning, and their goal is to learn a global model that follows general instructions rather than to personalize the LLM's behavior for each client. In this paper, we consider two datasets for personalized LLM: \textbf{Task-heterogeneous instruction following:} Following \cite{yang2024dualpersonal}, we randomly select $8$ NLP tasks from Flan \cite{wei2021finetuned}. One task per client, resulting in $8$ total clients. Each task randomly selects $300$ training samples and $200$ test samples. ROUGE-1 is used as the evaluation metric. \textbf{Persona-aligned instruction following:} We also provide a more realistic scenario with RoleBench \cite{wang2023rolellm}, where each client takes on a specific role and has a Q\&A dataset with clear role characteristics. Details can be found in Appendix~\ref{sec llmdataset}.

\textbf{System Settings} We adapt Alpaca-LoRA\footnote{\url{https://github.com/tloen/alpaca-lora}} as the base model, initialized with LLaMA3-8B \cite{dubey2024llama3}. The rank of LoRA is set to $8$. We employ a linear learning rate decay schedule from $5e{-}5$ to $5e{-}6$ over the training process. The number of communication rounds is set to $50$ to ensure convergence. For each round, the local training is conducted with $20$ steps, and the batch size is set to $4$.

\textbf{Parameter Scale} Similar to classic FL settings, we vary the number of LoRA modules for Multi-model methods and vary the rank of LoRA modules for Single-model methods. Details are shown in Table~\ref{tab:merged_param_scale}. For cluster FL methods like IFCA, the number of global models is limited to at most $8$, equal to the number of clients. All experiments are conducted on a single NVIDIA A40 GPU.

\begin{figure}[t]
    \centering
    \includegraphics[width=0.95\linewidth]{figures/flan_curve.pdf}
    \caption{Server parameter usage vs. accuracy on the federated Flan dataset. The values on the x-axis are shown as multiples of the parameter count of a Rank-8 LoRA. The highest and second-highest accuracy points are marked with black and gray respectively.}
    \label{fig:server_flan}
\end{figure}

Following Table~\ref{tab:merged_param_scale}, we vary Single-model methods by the LoRA rank and Multi-model methods by the number of LoRA modules. The server parameter usage vs. accuracy for the federated Flan dataset is shown in Fig.~\ref{fig:server_flan}. \ours demonstrates better instruction-following abilities on the federated Flan dataset by sharing up to $12$ LoRA modules globally while only transmitting one LoRA module of rank $8$ per client.  Results for the federated RoleBench dataset can be found in Appendix~\ref{result_character_llm}.

\textbf{Model Merging (\ours) provides a valuable alternative to MoE-like methods (FedEM) for FL with foundation models while being more client-resource friendly.} Unlike classic settings, MoE-like methods are widely used in foundation models. A typical setup for a Mixture-of-Experts (MoE) foundation model in a federated learning scenario is that each client hosts several experts and a local router. These experts are sent to the server and aggregated separately to form a pool of global experts. While being simple in optimization strategy, it is not as client-resource-efficient as \ours. A detailed comparison between FedEM and \ours is presented in Table~\ref{tab:fedem_fedmerge_taskwise}. We select the best performance for FedEM and \ours within the rank settings in Table~\ref{tab:merged_param_scale}, which is $12$ shared LoRA modules for FedEM and $8$ for \ours. From Table~\ref{tab:fedem_fedmerge_taskwise}, \ours and FedEM outperform each other on different tasks, while \ours is better in average performance across clients.

\section{Conclusion}

In this paper, we propose an end-to-end Model Merging framework for Federated Learning, where merging weights and global models are optimized jointly from scratch. The server manages all global models and merging weights, customizing personalized models for each client by weight-averaging the global models. \ours can boost performance with a large model soup on the server, while still being client-resource friendly by only communicating and optimizing a single model for each client. \ours offers a promising direction for Federated Foundation Models: a large server knowledge pool can match foundation model scales, while each client selects and combines a small subset of parameters without transmitting the full model.

\newpage
\bibliography{aaai2026}

@article{yi2024fedmoe,
  title={FedMoE: Data-Level Personalization with Mixture of Experts for Model-Heterogeneous Personalized Federated Learning},
  author={Yi, Liping and Yu, Han and Ren, Chao and Zhang, Heng and Wang, Gang and Liu, Xiaoguang and Li, Xiaoxiao},
  journal={arXiv preprint arXiv:2402.01350},
  year={2024}
}

@article{long2024dual,
  title={Dual-personalizing adapter for federated foundation models},
  author={Long, Guodong and Shen, Tao and Jiang, Jing and Blumenstein, Michael and others},
  journal={Advances in Neural Information Processing Systems},
  volume={37},
  pages={39409--39433},
  year={2024}
}

@article{krizhevsky2009learningcifar100,
  title={Learning multiple layers of features from tiny images},
  author={Krizhevsky, Alex and Hinton, Geoffrey and others},
  year={2009},
  publisher={Toronto, ON, Canada}
}

@inproceedings{bhuyan2022multi,
  title={Multi-model federated learning},
  author={Bhuyan, Neelkamal and Moharir, Sharayu},
  booktitle={2022 14th International Conference on COMmunication Systems \& NETworkS (COMSNETS)},
  pages={779--783},
  year={2022},
  organization={IEEE}
}

@article{dubey2024llama3,
  title={The llama 3 herd of models},
  author={Dubey, Abhimanyu and Jauhri, Abhinav and Pandey, Abhinav and Kadian, Abhishek and Al-Dahle, Ahmad and Letman, Aiesha and Mathur, Akhil and Schelten, Alan and Yang, Amy and Fan, Angela and others},
  journal={arXiv preprint arXiv:2407.21783},
  year={2024}
}

@inproceedings{zhang2024towardsllmfedavg,
  title={Towards building the federatedGPT: Federated instruction tuning},
  author={Zhang, Jianyi and Vahidian, Saeed and Kuo, Martin and Li, Chunyuan and Zhang, Ruiyi and Yu, Tong and Wang, Guoyin and Chen, Yiran},
  booktitle={ICASSP 2024-2024 IEEE International Conference on Acoustics, Speech and Signal Processing (ICASSP)},
  pages={6915--6919},
  year={2024},
  organization={IEEE}
}

@article{wang2025never,
  title={Never Start from Scratch: Expediting On-Device LLM Personalization via Explainable Model Selection},
  author={Wang, Haoming and Yang, Boyuan and Yin, Xiangyu and Gao, Wei},
  journal={arXiv preprint arXiv:2504.13938},
  year={2025}
}

@article{kim2025personalized,
  title={Personalized Language Models via Privacy-Preserving Evolutionary Model Merging},
  author={Kim, Kyuyoung and Shin, Jinwoo and Kim, Jaehyung},
  journal={arXiv preprint arXiv:2503.18008},
  year={2025}
}

@article{yi2023fedlora,
  title={Fedlora: Model-heterogeneous personalized federated learning with lora tuning},
  author={Yi, Liping and Yu, Han and Wang, Gang and Liu, Xiaoguang},
  journal={arXiv preprint arXiv:2310.13283},
  year={2023}
}

@inproceedings{li2017deeper,
  title={Deeper, broader and artier domain generalization},
  author={Li, Da and Yang, Yongxin and Song, Yi-Zhe and Hospedales, Timothy M},
  booktitle={Proceedings of the IEEE international conference on computer vision},
  pages={5542--5550},
  year={2017}
}

@article{hsu2019measuringnon-iid,
  title={Measuring the effects of non-identical data distribution for federated visual classification},
  author={Hsu, Tzu-Ming Harry and Qi, Hang and Brown, Matthew},
  journal={arXiv preprint arXiv:1909.06335},
  year={2019}
}

@article{yadav2024survey,
  title={A survey on model moerging: Recycling and routing among specialized experts for collaborative learning},
  author={Yadav, Prateek and Raffel, Colin and Muqeeth, Mohammed and Caccia, Lucas and Liu, Haokun and Chen, Tianlong and Bansal, Mohit and Choshen, Leshem and Sordoni, Alessandro},
  journal={arXiv preprint arXiv:2408.07057},
  year={2024}
}

@article{yang2024model,
  title={Model merging in llms, mllms, and beyond: Methods, theories, applications and opportunities},
  author={Yang, Enneng and Shen, Li and Guo, Guibing and Wang, Xingwei and Cao, Xiaochun and Zhang, Jie and Tao, Dacheng},
  journal={arXiv preprint arXiv:2408.07666},
  year={2024}
}

@article{lorahubhuang2023lorahub,
  title={Lorahub: Efficient cross-task generalization via dynamic lora composition},
  author={Huang, Chengsong and Liu, Qian and Lin, Bill Yuchen and Pang, Tianyu and Du, Chao and Lin, Min},
  journal={arXiv preprint arXiv:2307.13269},
  year={2023}
}

@article{taskvectorilharco2022editing,
  title={Editing models with task arithmetic},
  author={Ilharco, Gabriel and Ribeiro, Marco Tulio and Wortsman, Mitchell and Gururangan, Suchin and Schmidt, Ludwig and Hajishirzi, Hannaneh and Farhadi, Ali},
  journal={arXiv preprint arXiv:2212.04089},
  year={2022}
}

@article{fishermergingmatena2022merging,
  title={Merging models with fisher-weighted averaging},
  author={Matena, Michael S and Raffel, Colin A},
  journal={Advances in Neural Information Processing Systems},
  volume={35},
  pages={17703--17716},
  year={2022}
}

@article{ostapenko2024towards,
  title={Towards modular llms by building and reusing a library of loras},
  author={Ostapenko, Oleksiy and Su, Zhan and Ponti, Edoardo Maria and Charlin, Laurent and Roux, Nicolas Le and Pereira, Matheus and Caccia, Lucas and Sordoni, Alessandro},
  journal={arXiv preprint arXiv:2405.11157},
  year={2024}
}

@article{yosinski2014transferable,
  title={How transferable are features in deep neural networks?},
  author={Yosinski, Jason and Clune, Jeff and Bengio, Yoshua and Lipson, Hod},
  journal={Advances in neural information processing systems},
  volume={27},
  year={2014}
}

@article{shi2024culturebank,
  title={Culturebank: An online community-driven knowledge base towards culturally aware language technologies},
  author={Shi, Weiyan and Li, Ryan and Zhang, Yutong and Ziems, Caleb and Horesh, Raya and de Paula, Rog{\'e}rio Abreu and Yang, Diyi and others},
  journal={arXiv preprint arXiv:2404.15238},
  year={2024}
}

@article{wei2021finetuned,
  title={Finetuned language models are zero-shot learners},
  author={Wei, Jason and Bosma, Maarten and Zhao, Vincent Y and Guu, Kelvin and Yu, Adams Wei and Lester, Brian and Du, Nan and Dai, Andrew M and Le, Quoc V},
  journal={arXiv preprint arXiv:2109.01652},
  year={2021}
}

@article{shao2023character,
  title={Character-llm: A trainable agent for role-playing},
  author={Shao, Yunfan and Li, Linyang and Dai, Junqi and Qiu, Xipeng},
  journal={arXiv preprint arXiv:2310.10158},
  year={2023}
}

@article{wang2023rolellm,
  title={Rolellm: Benchmarking, eliciting, and enhancing role-playing abilities of large language models},
  author={Wang, Zekun Moore and Peng, Zhongyuan and Que, Haoran and Liu, Jiaheng and Zhou, Wangchunshu and Wu, Yuhan and Guo, Hongcheng and Gan, Ruitong and Ni, Zehao and Yang, Jian and others},
  journal={arXiv preprint arXiv:2310.00746},
  year={2023}
}

@article{hu2022text,
  title={Text style transfer: A review and experimental evaluation},
  author={Hu, Zhiqiang and Lee, Roy Ka-Wei and Aggarwal, Charu C and Zhang, Aston},
  journal={ACM SIGKDD Explorations Newsletter},
  volume={24},
  number={1},
  pages={14--45},
  year={2022},
  publisher={ACM New York, NY, USA}
}

@article{fisher2024styleremix,
  title={StyleRemix: Interpretable Authorship Obfuscation via Distillation and Perturbation of Style Elements},
  author={Fisher, Jillian and Hallinan, Skyler and Lu, Ximing and Gordon, Mitchell and Harchaoui, Zaid and Choi, Yejin},
  journal={arXiv preprint arXiv:2408.15666},
  year={2024}
}

@article{li2024culturellm,
  title={Culturellm: Incorporating cultural differences into large language models},
  author={Li, Cheng and Chen, Mengzhuo and Wang, Jindong and Sitaram, Sunayana and Xie, Xing},
  journal={Advances in Neural Information Processing Systems},
  volume={37},
  pages={84799--84838},
  year={2024}
}

@inproceedings{kuang2024federatedscope,
  title={Federatedscope-llm: A comprehensive package for fine-tuning large language models in federated learning},
  author={Kuang, Weirui and Qian, Bingchen and Li, Zitao and Chen, Daoyuan and Gao, Dawei and Pan, Xuchen and Xie, Yuexiang and Li, Yaliang and Ding, Bolin and Zhou, Jingren},
  booktitle={Proceedings of the 30th ACM SIGKDD Conference on Knowledge Discovery and Data Mining},
  pages={5260--5271},
  year={2024}
}

@article{ye2024fedllm,
  title={Fedllm-bench: Realistic benchmarks for federated learning of large language models},
  author={Ye, Rui and Ge, Rui and Zhu, Xinyu and Chai, Jingyi and Yaxin, Du and Liu, Yang and Wang, Yanfeng and Chen, Siheng},
  journal={Advances in Neural Information Processing Systems},
  volume={37},
  pages={111106--111130},
  year={2024}
}

@inproceedings{ye2024openfedllm,
  title={Openfedllm: Training large language models on decentralized private data via federated learning},
  author={Ye, Rui and Wang, Wenhao and Chai, Jingyi and Li, Dihan and Li, Zexi and Xu, Yinda and Du, Yaxin and Wang, Yanfeng and Chen, Siheng},
  booktitle={Proceedings of the 30th ACM SIGKDD conference on knowledge discovery and data mining},
  pages={6137--6147},
  year={2024}
}

@article{rumelhart1986learning,
  title={Learning representations by back-propagating errors},
  author={Rumelhart, David E and Hinton, Geoffrey E and Williams, Ronald J},
  journal={nature},
  volume={323},
  number={6088},
  pages={533--536},
  year={1986},
  publisher={Nature Publishing Group UK London}
}

@inproceedings{li2021ditto,
  title={Ditto: Fair and robust federated learning through personalization},
  author={Li, Tian and Hu, Shengyuan and Beirami, Ahmad and Smith, Virginia},
  booktitle={International Conference on Machine Learning},
  pages={6357--6368},
  year={2021},
  organization={PMLR}
}

@article{t2020personalized,
  title={Personalized federated learning with moreau envelopes},
  author={T Dinh, Canh and Tran, Nguyen and Nguyen, Josh},
  journal={Advances in Neural Information Processing Systems},
  volume={33},
  pages={21394--21405},
  year={2020}
}

@article{chen2024local,
  title={Local Superior Soups: A Catalyst for Model Merging in Cross-Silo Federated Learning},
  author={Chen, Minghui and Jiang, Meirui and Zhang, Xin and Dou, Qi and Wang, Zehua and Li, Xiaoxiao},
  journal={arXiv preprint arXiv:2410.23660},
  year={2024}
}

@article{fedemmarfoq2021federated,
  title={Federated multi-task learning under a mixture of distributions},
  author={Marfoq, Othmane and Neglia, Giovanni and Bellet, Aur{\'e}lien and Kameni, Laetitia and Vidal, Richard},
  journal={Advances in Neural Information Processing Systems},
  volume={34},
  pages={15434--15447},
  year={2021}
}

@inproceedings{collins2021exploiting,
  title={Exploiting shared representations for personalized federated learning},
  author={Collins, Liam and Hassani, Hamed and Mokhtari, Aryan and Shakkottai, Sanjay},
  booktitle={International conference on machine learning},
  pages={2089--2099},
  year={2021},
  organization={PMLR}
}

@article{long2023multi,
  title={Multi-center federated learning: clients clustering for better personalization},
  author={Long, Guodong and Xie, Ming and Shen, Tao and Zhou, Tianyi and Wang, Xianzhi and Jiang, Jing},
  journal={World Wide Web},
  volume={26},
  number={1},
  pages={481--500},
  year={2023},
  publisher={Springer}
}

@article{yang2024dualpersonal,
  title={Dual-Personalizing Adapter for Federated Foundation Models},
  author={Yang, Yiyuan and Long, Guodong and Shen, Tao and Jiang, Jing and Blumenstein, Michael},
  journal={arXiv preprint arXiv:2403.19211},
  year={2024}
}

@inproceedings{wu2023personalizedgaussianmixture,
  title={Personalized federated learning under mixture of distributions},
  author={Wu, Yue and Zhang, Shuaicheng and Yu, Wenchao and Liu, Yanchi and Gu, Quanquan and Zhou, Dawei and Chen, Haifeng and Cheng, Wei},
  booktitle={International Conference on Machine Learning},
  pages={37860--37879},
  year={2023},
  organization={PMLR}
}

@inproceedings{mcmahan2017communication,
  title={Communication-efficient learning of deep networks from decentralized data},
  author={McMahan, Brendan and Moore, Eider and Ramage, Daniel and Hampson, Seth and y Arcas, Blaise Aguera},
  booktitle={Artificial intelligence and statistics},
  pages={1273--1282},
  year={2017},
  organization={PMLR}
}

@article{ghosh2020efficientifca,
  title={An efficient framework for clustered federated learning},
  author={Ghosh, Avishek and Chung, Jichan and Yin, Dong and Ramchandran, Kannan},
  journal={Advances in Neural Information Processing Systems},
  volume={33},
  pages={19586--19597},
  year={2020}
}

@inproceedings{cho2024heterogeneouslora,
  title={Heterogeneous lora for federated fine-tuning of on-device foundation models},
  author={Cho, Yae Jee and Liu, Luyang and Xu, Zheng and Fahrezi, Aldi and Joshi, Gauri},
  booktitle={Proceedings of the 2024 Conference on Empirical Methods in Natural Language Processing},
  pages={12903--12913},
  year={2024}
}

@article{sun2024improvingloraprivacy,
  title={Improving loRA in privacy-preserving federated learning},
  author={Sun, Youbang and Li, Zitao and Li, Yaliang and Ding, Bolin},
  journal={arXiv preprint arXiv:2403.12313},
  year={2024}
}

@inproceedings{modelsoupwortsman2022model,
  title={Model soups: averaging weights of multiple fine-tuned models improves accuracy without increasing inference time},
  author={Wortsman, Mitchell and Ilharco, Gabriel and Gadre, Samir Ya and Roelofs, Rebecca and Gontijo-Lopes, Raphael and Morcos, Ari S and Namkoong, Hongseok and Farhadi, Ali and Carmon, Yair and Kornblith, Simon and others},
  booktitle={International conference on machine learning},
  pages={23965--23998},
  year={2022},
  organization={PMLR}
}

@article{lu2023routing,
  title={Routing to the expert: Efficient reward-guided ensemble of large language models},
  author={Lu, Keming and Yuan, Hongyi and Lin, Runji and Lin, Junyang and Yuan, Zheng and Zhou, Chang and Zhou, Jingren},
  journal={arXiv preprint arXiv:2311.08692},
  year={2023}
}

@article{loraflowwang2024lora,
  title={LoRA-Flow: Dynamic LoRA Fusion for Large Language Models in Generative Tasks},
  author={Wang, Hanqing and Ping, Bowen and Wang, Shuo and Han, Xu and Chen, Yun and Liu, Zhiyuan and Sun, Maosong},
  journal={arXiv preprint arXiv:2402.11455},
  year={2024}
}

@article{chronopoulou2023adaptersoup,
  title={Adaptersoup: Weight averaging to improve generalization of pretrained language models},
  author={Chronopoulou, Alexandra and Peters, Matthew E and Fraser, Alexander and Dodge, Jesse},
  journal={arXiv preprint arXiv:2302.07027},
  year={2023}
}

@article{lu2024twin,
  title={Twin-merging: Dynamic integration of modular expertise in model merging},
  author={Lu, Zhenyi and Fan, Chenghao and Wei, Wei and Qu, Xiaoye and Chen, Dangyang and Cheng, Yu},
  journal={arXiv preprint arXiv:2406.15479},
  year={2024}
}

@misc{huggingface,
  author = {Hugging Face},
  title = {Hugging Face - The AI Community Building the Future},
  howpublished = {\url{https://huggingface.co}},
  year = {2023},
  note = {Accessed: October 15, 2023}
}

@article{zhao2024retrieval,
  title={Retrieval-augmented mixture of lora experts for uploadable machine learning},
  author={Zhao, Ziyu and Gan, Leilei and Wang, Guoyin and Hu, Yuwei and Shen, Tao and Yang, Hongxia and Kuang, Kun and Wu, Fei},
  journal={arXiv preprint arXiv:2406.16989},
  year={2024}
}

@inproceedings{feng2025pm,
  title={PM-MOE: Mixture of Experts on Private Model Parameters for Personalized Federated Learning},
  author={Feng, Yu and Geng, Yangli-ao and Zhu, Yifan and Han, Zongfu and Yu, Xie and Xue, Kaiwen and Luo, Haoran and Sun, Mengyang and Zhang, Guangwei and Song, Meina},
  booktitle={Proceedings of the ACM on Web Conference 2025},
  pages={134--146},
  year={2025}
}

@inproceedings{ijcai2025p610,
  title     = {Heterogeneous Federated Learning with Scalable Server Mixture-of-Experts},
  author    = {Jiang, Jingang and Chen, Yanzhao and Liu, Xiangyang and Jiang, Haiqi and Fan, Chenyou},
  booktitle = {Proceedings of the Thirty-Fourth International Joint Conference on
               Artificial Intelligence, {IJCAI-25}},
  publisher = {International Joint Conferences on Artificial Intelligence Organization},
  editor    = {James Kwok},
  pages     = {5480--5488},
  year      = {2025},
  month     = {8},
  note      = {Main Track},
  doi       = {10.24963/ijcai.2025/610},
  url       = {https://doi.org/10.24963/ijcai.2025/610},
}

@inproceedings{radwan2025feddg,
  title={FedDG-MoE: Test-Time Mixture-of-Experts Fusion for Federated Domain Generalization},
  author={Radwan, Ahmed and Soliman, Mahmoud and Abdelaziz, Omar and Shehata, Mohamed},
  booktitle={Proceedings of the Computer Vision and Pattern Recognition Conference},
  pages={1811--1820},
  year={2025}
}

@inproceedings{xie2025dflmoe,
  title={dFLMoE: Decentralized Federated Learning via Mixture of Experts for Medical Data Analysis},
  author={Xie, Luyuan and Luan, Tianyu and Cai, Wenyuan and Yan, Guochen and Chen, Zhaoyu and Xi, Nan and Fang, Yuejian and Shen, Qingni and Wu, Zhonghai and Yuan, Junsong},
  booktitle={Proceedings of the Computer Vision and Pattern Recognition Conference},
  pages={10203--10213},
  year={2025}
}

@article{luo2024mixture,
  title={Mixture of experts made personalized: Federated prompt learning for vision-language models},
  author={Luo, Jun and Chen, Chen and Wu, Shandong},
  journal={arXiv preprint arXiv:2410.10114},
  year={2024}
}

@article{tao2024task,
  title={Task arithmetic through the lens of one-shot federated learning},
  author={Tao, Zhixu Silvia and Mason, Ian and Kulkarni, Sanjeev and Boix, Xavier},
  journal={arXiv preprint arXiv:2411.18607},
  year={2024}
}

@article{salami2024closed,
  title={Closed-form merging of parameter-efficient modules for federated continual learning},
  author={Salami, Riccardo and Buzzega, Pietro and Mosconi, Matteo and Bonato, Jacopo and Sabetta, Luigi and Calderara, Simone},
  journal={arXiv preprint arXiv:2410.17961},
  year={2024}
}

@article{jin2022dataless,
  title={Dataless knowledge fusion by merging weights of language models},
  author={Jin, Xisen and Ren, Xiang and Preotiuc-Pietro, Daniel and Cheng, Pengxiang},
  journal={arXiv preprint arXiv:2212.09849},
  year={2022}
}

@article{chen2025breaking,
  title={Breaking the Aggregation Bottleneck in Federated Recommendation: A Personalized Model Merging Approach},
  author={Chen, Jundong and Zhang, Honglei and Zhang, Chunxu and Luo, Fangyuan and Li, Yidong},
  journal={arXiv preprint arXiv:2508.12386},
  year={2025}
}

@article{zhou2025mergeme,
  title={MergeME: Model merging techniques for homogeneous and heterogeneous MoEs},
  author={Zhou, Yuhang and Karamanolakis, Giannis and Soto, Victor and Rumshisky, Anna and Kulkarni, Mayank and Huang, Furong and Ai, Wei and Lu, Jianhua},
  journal={arXiv preprint arXiv:2502.00997},
  year={2025}
}

@article{cheng2025whoever,
  title={Whoever started the interference should end it: Guiding data-free model merging via task vectors},
  author={Cheng, Runxi and Xiong, Feng and Wei, Yongxian and Zhu, Wanyun and Yuan, Chun},
  journal={arXiv preprint arXiv:2503.08099},
  year={2025}
}

@article{liu2024fedlpa,
  title={Fedlpa: One-shot federated learning with layer-wise posterior aggregation},
  author={Liu, Xiang and Liu, Liangxi and Ye, Feiyang and Shen, Yunheng and Li, Xia and Jiang, Linshan and Li, Jialin},
  journal={Advances in Neural Information Processing Systems},
  volume={37},
  pages={81510--81548},
  year={2024}
}

@inproceedings{du2025adamms,
  title={Adamms: Model merging for heterogeneous multimodal large language models with unsupervised coefficient optimization},
  author={Du, Yiyang and Wang, Xiaochen and Chen, Chi and Ye, Jiabo and Wang, Yiru and Li, Peng and Yan, Ming and Zhang, Ji and Huang, Fei and Sui, Zhifang and others},
  booktitle={Proceedings of the Computer Vision and Pattern Recognition Conference},
  pages={9413--9422},
  year={2025}
}

\clearpage
\appendix

\section{More Experiments on Classic FL}
\label{sec:pacs}
\subsection{Client Parameter Usage vs. Performance}
\label{sec puvsacc}
From Fig.~\ref{personal_curve_share}, the performance of single-model methods degrades as model scale increases. This is because both Personal and Cluster Non-IID settings involve 50 clients, where the large number of clients with sparse client data limits performance gains from scaling up the model. Therefore, we build Non-IID setting with less heterogeneity. Each client is assigned sufficient number of samples to ensure scaling up model architectures more effective:

\textbf{Domain Non-IID}: We use the PACS \cite{li2017deeper} dataset, assigning each client to one domain, resulting in 4 clients.  Note that we do not run IFCA on Domain Non-IID since having only 4 clients is too few for Cluster FL methods.

\begin{figure}[t]
    \centering
    \includegraphics[width=0.95\linewidth]{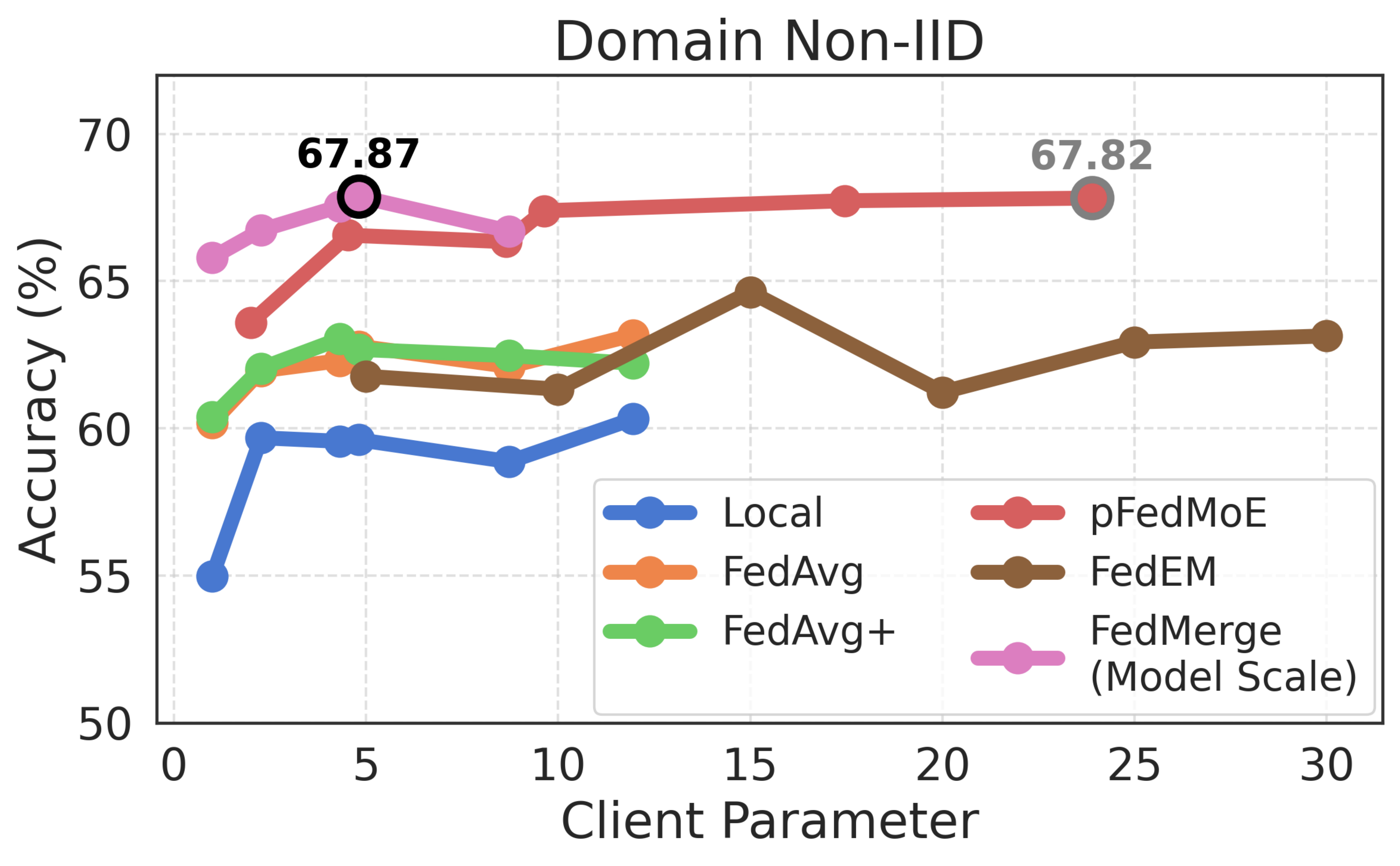}
    \caption{Client parameter usage vs. accuracy. The x-axis represents multiples of ResNet-9's parameter count. In FedMerge, we fix the number of global models at 15, all using the same architecture, selected from [ResNet-9, ResNet-18, ResNet-34, ResNet-50, and ResNet-101].}
    \label{fig:domain non-IID}
\end{figure}

In Fig.~\ref{fig:domain non-IID}, we present client parameter usage versus accuracy for Domain Non-IID. As shown, the performance of single-model methods improves significantly as model scale increases.

\textbf{FedMerge further enhances performance through model scaling while remaining client resource-efficient.} In Fig.~\ref{fig:domain non-IID}, we evaluate \ours with a fixed set of 15 global models while varying the model architecture across [ResNet-9, ResNet-18, ResNet-34, ResNet-50, and ResNet-101]. Since \ours requires only a single model to be transmitted and optimized per client, it remains highly resource-efficient even as model scale increases. As demonstrated in Fig.~\ref{fig:domain non-IID}, \ours outperforms pFedMoE by leveraging model scaling while incurring lower client-side resource consumption.

\subsection{More Discussion on Merging Weight Matrix}
\begin{figure*}[ht]
    \centering
    \begin{subfigure}[b]{0.98\textwidth}
        \centering
        \includegraphics[width=\linewidth]{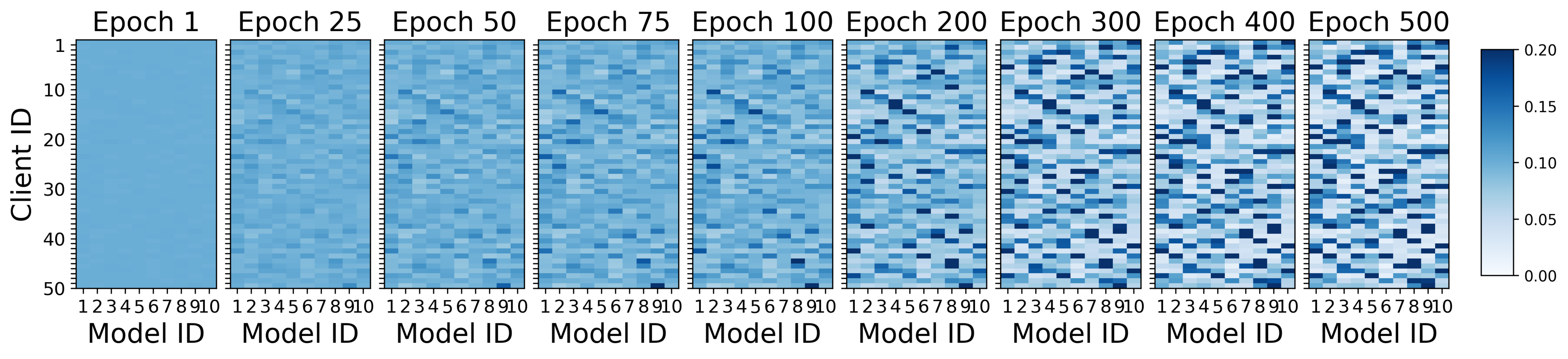}
        \caption{Heat map of merge weights for Personal Non-IID.}
        \label{fig:app_sub1}
    \end{subfigure}
    \hfill
    \begin{subfigure}[b]{0.98\textwidth}
        \centering
        \includegraphics[width=\linewidth]{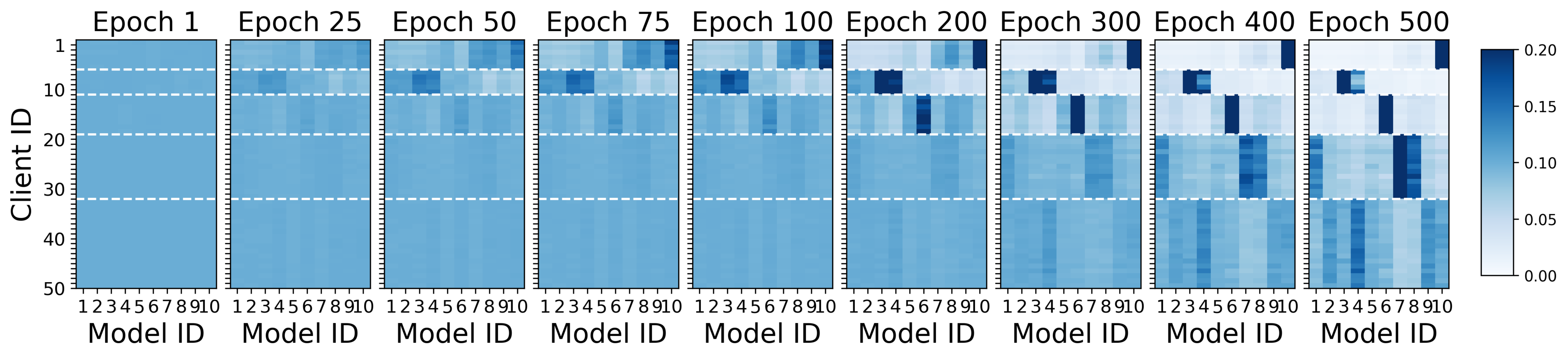}
        \caption{Heat map of merge weights for Cluster Non-IID.}
        \label{fig:app W strcutre cluster}
    \end{subfigure}

    \caption{Additional visualizations of merge weights with 10 global models and 50 clients. For Cluster Non-IID, the convergence speed varies across clusters.}
    \label{fig:app W strcutre}
\end{figure*}

We provide additional visualizations of the merging weight matrix in Fig.~\ref{fig:app W strcutre}. Specifically, in Fig.~\ref{fig:app W strcutre cluster}, the evolution speed of different clusters varies. Smaller clusters with fewer clients evolve faster and demonstrate distinct patterns within 200 epochs, as observed in the first three clusters with small client IDs. However, for the largest cluster (clients with IDs [32, ..., 50]), the merging weights evolve more slowly, even after 500 epochs. This is because the largest cluster has the highest number of clients (19 clients) and requires more global models, resulting in a more uniform merging weight distribution. This does not mean that large clusters are difficult to converge. It simply indicates that the weight updates are more evenly distributed within larger clusters. These small weight changes within large clusters are sufficient to significantly distinguish large clusters from other clusters.

\section{More Experiments on FL with Foundation Models}
\subsection{More Discussion on Personalized LLM Datasets} 
\label{sec llmdataset}

\textbf{Dataset for Persona-Aligned Instruction Following.} There exist several baselines \cite{ye2024openfedllm,kuang2024federatedscope,ye2024fedllm} for federated foundation models. However, these models are primarily designed for instruction fine-tuning, and their goal is to learn a global model to follow general instructions rather than personalize the LLM's behavior to each client. The evaluation metrics in these baselines are often datasets or evaluation models independent of the FL training environment. Therefore, to develop personalized federated LLMs, it is essential to focus on datasets that reflect individual characteristics—such as culture \cite{li2024culturellm,shi2024culturebank}, style \cite{fisher2024styleremix,hu2022text}, and personality traits \cite{wang2023rolellm,shao2023character}. In this paper, we use RoleBench \cite{wang2023rolellm}, a collection of $100$ characters. The role-specific instructions and responses are used as the personalized dataset for each client. Each client is assigned one specific character, and we randomly sample $50$ characters out of $100$, resulting in $50$ clients. For each selected character, we randomly select half of the original training dataset and test dataset as the client-specific dataset. Table~\ref{tab:roleplay_combined} presents some input-output examples. Following the evaluation metric in \cite{wang2023rolellm}, ROUGE-L is used as the evaluation metric. At each communication round, $20$ clients are randomly sampled out of $50$.

\begin{table}[htbp]
\centering
\small
\resizebox{0.95\linewidth}{!}{%
\begin{tabular}{ll}
\toprule
\textbf{Task} & \textbf{Dataset} \\
\midrule
Paraphrase & glue\_qqp \\
Entailment & snli \\
Structure to text & web\_nlg\_en \\
Text formatting & fix\_punct \\
Linguistic acceptability & cola \\
Word disambiguation & wic \\
Coreference & definite\_pronoun\_resolution \\
Question classification & trec \\
\bottomrule
\end{tabular}
}

\caption{Federated Flan dataset for personalized instruction Tuning.}
\label{tab:federated_dataset1}
\end{table}

\textbf{Dataset for Task-Heterogeneous Instruction Following.} We construct a federated dataset from Flan \cite{wei2021finetuned}, which is a collection of various classic NLP tasks from over 60 datasets for instruction tuning. Following \cite{long2024dual}, we randomly select $8$ NLP tasks, one task for one client, resulting in $8$ total clients. For each task, we randomly select $300$ training samples and $200$ test samples from the original dataset. ROUGE-1 is used as the evaluation metric. We perform full client selection.

\begin{table}[htbp]
\centering
\small
\renewcommand{\arraystretch}{1.3}
\begin{tabular}{@{}p{0.25\linewidth}p{0.72\linewidth}@{}}
\specialrule{1pt}{0pt}{1pt} 
\textbf{Role} & Jack Sparrow — A witty and eccentric pirate captain from the \textit{Pirates of the Caribbean} film series. \\
\midrule
\textbf{Question} & How did you manage to escape from the casket? \\
\textbf{Answer} & Ah, well, you see, I always have a few tricks up my sleeve, or in this case, up my coffin. It wasn't easy, though, as I had to wait for just the right moment to break open the casket without drawing too much attention to myself. But, as they say, desperate times call for desperate measures, and I've always been good at thinking on my feet. \\
\specialrule{1pt}{1pt}{1pt} 
\textbf{Role} & Stephen Hawking — A renowned theoretical physicist known for his work on black holes and cosmology. \\
\midrule
\textbf{Question} & What motivated you to pursue a career in cosmology? \\
\textbf{Answer} & I've always been fascinated by the mysteries of the universe, and I wanted to dedicate my life to uncovering its secrets. I believe that through science, we can better understand our place in the world and how everything is interconnected. \\
\specialrule{1pt}{1pt}{0pt} 
\end{tabular}

\caption{Roleplay examples in the federated RoleBench dataset, showing character-specific questions and responses. Each question is an input prompt to an LLM, and the corresponding answer is the generated response.}
\label{tab:roleplay_combined}

\end{table}

\subsection{Baseline Settings for Foundation-model FL Experiments}
\label{baseline_llm}

\textbf{pFedMoE} \cite{yi2024fedmoe} employs a personalized router to balance contributions from the global and local models. Following this design, we introduce a router for each pair of global and local LoRA modules. Each router is implemented as a linear layer followed by a softmax activation. The router takes as input the same embedding used by the LoRA modules and outputs a routing score. This score is then used to compute a weighted average of the outputs from the global and local LoRA modules, yielding the final LoRA output.

\begin{table*}[t]
\centering
\caption{Parameter scale settings for LoRA modules on the Federated RoleBench dataset. Parameter scale is measured relative to Rank-8.}
\fontsize{9pt}{10pt}\selectfont
\resizebox{\textwidth}{!}{%
\begin{tabular}{@{}llcccccc@{}}
\toprule
\multirow{2}{*}{\textbf{Multi-Model}} 
& \textbf{Description}         & Rank-8×5 & Rank-8×10 & Rank-8×15 & Rank-8×20 & Rank-8×25 & Rank-8×30 \\
& \textbf{Parameter Scale}     & 5        & 10        & 15        & 20        & 25        & 30        \\
\midrule
\multirow{2}{*}{\textbf{Single-Model}} 
& \textbf{Description}         & Rank-8   & Rank-16   & Rank-24   & Rank-32   &           &           \\
& \textbf{Parameter Scale}     & 1        & 2         & 3         & 4         &           &           \\
\bottomrule
\end{tabular}%
}
\label{lora_param_scale_rolebench}
\end{table*}

\textbf{FedEM} \cite{fedemmarfoq2021federated} requires a personalized router for a group of global models. Similar to pFedMoE, we add a router for each set of global LoRAs, and the output dimension of the router is the number of global LoRAs. The input embedding of the router is also the input to the LoRA module. The router produces routing scores, which are then used to compute a weighted combination of the global LoRA outputs, yielding the final LoRA output.

\begin{figure}[t]
    \centering
    \includegraphics[width=0.95\linewidth]{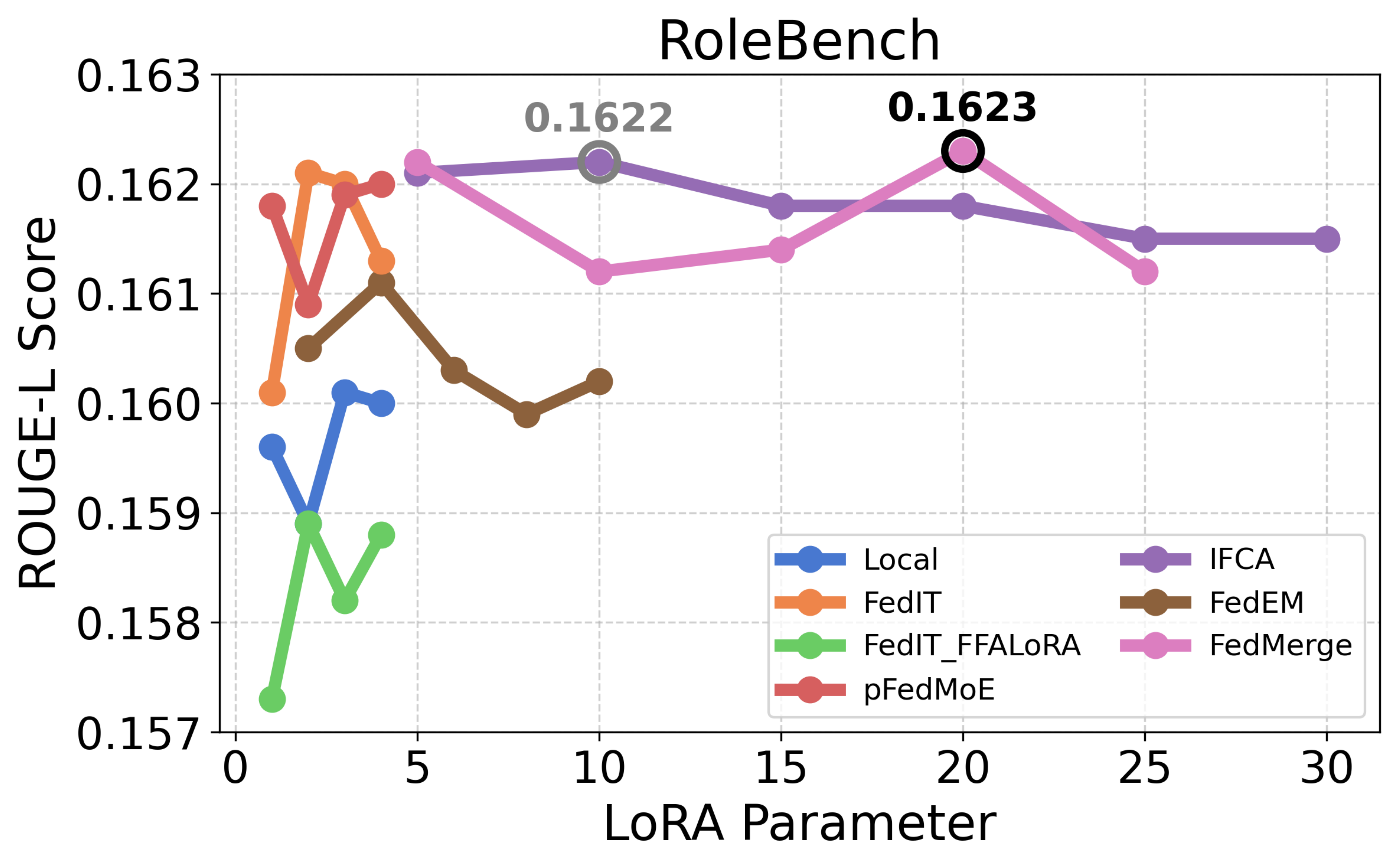}
    \caption{Server parameter usage vs. accuracy on the federated RoleBench dataset. The values on the x-axis are shown as multiples of the parameter count of a Rank-8 LoRA. The highest and second-highest accuracy points are marked in black and gray, respectively.}
    \label{fig:server_character}
\end{figure}

\textbf{FedIT-FFALoRA} As in \cite{sun2024improvingloraprivacy}, we only transmit and optimize the $B$ matrix of LoRA. The $A$ matrix is initialized and frozen on each client.

\textbf{FedMerge} Due to the varying lengths of predicted texts across clients, the gradient magnitudes returned to the server differ significantly. As a result, the merging weights are updated at uneven rates, with some parameters receiving large updates while others remain almost unchanged. To address this non-IID issue, we normalize the gradient of each client's corresponding merging weight, ensuring that all clients contribute gradients with consistent magnitudes while preserving their unique update directions. We set the learning rate to 0.01, consistent with the learning rate used in classical FL settings.

\subsection{Results for Federated RoleBench Dataset}
\label{result_character_llm}

Following Table~\ref{lora_param_scale_rolebench}, we vary Single-Model methods by LoRA rank and Multi-Model methods by the number of LoRA modules. FedEM has to optimize all global LoRA modules simultaneously. Due to our limited computational resources, we only vary the number of shared LoRA modules for FedEM from $2$ to $10$. Since there are $50$ clients, we compute the average accuracy over the bottom-half clients to evaluate each method's ability to handle the worst-performing characteristics. The server parameter usage vs. accuracy for the federated RoleBench dataset is shown in Fig.~\ref{fig:server_character}. \ours outperforms baselines, demonstrating its effectiveness in supporting the worst-performing clients.

\section{Gradient Derivations for $\Theta$ (Eq~\eqref{eq:gradient_theta}) and $w$ (Eq~\eqref{eq:gradient_w})}
\label{sec:gradient_derivations}

Given the objective function:
\[
L(\{\Theta_j\}, \{w_{(i,j)}\}) = \sum_{i=1}^m \ell\left(Y_i, f(X_i, \theta_i)\right),
\]
where
\[
\theta_i = \sum_{j=1}^d w_{(i,j)} \Theta_j.
\]

The goal is to derive gradient for $\Theta$ and $w$ with $\theta$ as intermediate variable.

\subsection{Gradient with respect to \(\Theta_j\)}

1. Compute \(\frac{\partial \theta_i}{\partial \Theta_j}\):
\[
\frac{\partial \theta_i}{\partial \Theta_j} = w_{(i,j)}.
\]

2. Use the chain rule to compute \(\frac{\partial L}{\partial \Theta_j}\):
\[
\frac{\partial L}{\partial \Theta_j} = \sum_{i=1}^m \frac{\partial \ell\left(Y_i, f(X_i, \theta_i)\right)}{\partial \theta_i} \cdot \frac{\partial \theta_i}{\partial \Theta_j}.
\]

3. Substitute \(\frac{\partial \theta_i}{\partial \Theta_j}\):
\[
\frac{\partial L}{\partial \Theta_j} = \sum_{i=1}^m w_{(i,j)} \cdot \frac{\partial \ell\left(Y_i, f(X_i, \theta_i)\right)}{\partial \theta_i}.
\]

\subsection{Gradient with respect to \(w_{(i,j)}\)}

1. Compute \(\frac{\partial \theta_i}{\partial w_{(i,j)}}\):
\[
\frac{\partial \theta_i}{\partial w_{(i,j)}} = \Theta_j.
\]

2. Use the chain rule to compute \(\frac{\partial L}{\partial w_{(i,j)}}\):
\[
\frac{\partial L}{\partial w_{(i,j)}} = \frac{\partial \ell\left(Y_i, f(X_i, \theta_i)\right)}{\partial \theta_i} \cdot \frac{\partial \theta_i}{\partial w_{(i,j)}}.
\]

3. Substitute \(\frac{\partial \theta_i}{\partial w_{(i,j)}}\):
\[
\frac{\partial L}{\partial w_{(i,j)}} = \langle \Theta_j , \frac{\partial \ell\left(Y_i, f(X_i, \theta_i)\right)}{\partial \theta_i}\rangle.
\]

\subsection{Final Expressions}

Gradient with respect to \(\Theta_j\):
\[
\frac{\partial L}{\partial \Theta_j} = \sum_{i=1}^m w_{(i,j)} \cdot \frac{\partial \ell}{\partial \theta_i}.
\]

Gradient with respect to \(w_{(i,j)}\):\
\[
\frac{\partial L}{\partial w_{(i,j)}} = \langle \Theta_j , \frac{\partial \ell}{\partial \theta_i}\rangle.
\]

These expressions use \(\theta\) as an intermediate variable, facilitating efficient computation during implementation.

\section{Gradient Derivations for $\Theta$ and $w$ with Softmax Constraint}
\label{sec:softmax_gradient_derivations}

For clearer notation, we write the full merging weights as a weight matrix:

\[
\mathbf{w} =
\begin{bmatrix}
w_{1,1} & w_{1,2} & \cdots & w_{1,d} \\
w_{2,1} & w_{2,2} & \cdots & w_{2,d} \\
\vdots  & \vdots  & \ddots & \vdots  \\
w_{m,1} & w_{m,2} & \cdots & w_{m,d}
\end{bmatrix}
\]

where each row represents the personalized merging weights for each client. In the original form of the objective in Eq~\eqref{eq:MGFL_obj} and Eq~\eqref{eq:MGFL_constraint}, \(w\) is unconstrained and can take arbitrary values. In practice, this can result in an unstable training process as well as meaningless merging operations. For example, when the global models are neural networks, it is necessary to regularize the parameters before and after merging to prevent the parameters from becoming excessively large. 

We introduce a row-wise softmax function for \(w\) to ensure each element is non-negative and the sum of each row in \(w\) equals \(1\). Formally, \(w_{(i,j)}\) is obtained via a softmax function over parameters \(a_{(i,j)}\):
\[
w_{(i,j)} = \frac{\exp(a_{(i,j)})}{\sum_{j=1}^d \exp(a_{(i,j)})}.
\]

This ensures that each \(w_{(i,j)}\) satisfies \(0 \leq w_{(i,j)} \leq 1\) and \(\sum_{j=1}^d w_{(i,j)} = 1\) for each \(i\). Next we give out the gradient with respect to $a$.

\subsection{Gradient with respect to \(\Theta_j\)}

The gradient of \(L\) with respect to \(\Theta_j\) remains similar to the original derivation because \(\Theta_j\) appears directly in \(\theta_i\):
\[
\dfrac{\partial L}{\partial \Theta_j} = \sum_{i=1}^m w_{(i,j)} \cdot \dfrac{\partial \ell}{\partial \theta_i}.
\]

\subsection{Gradient with respect to \(a_{(i,j)}\)}

Since \(w_{(i,j)}\) now depends on \(a_{(i,j)}\) via the softmax function, we need to compute the gradient of \(L\) with respect to \(a_{(i,j)}\).

1. Compute \(\dfrac{\partial w_{(i,k)}}{\partial a_{(i,j)}}\):

The derivative of the softmax function is:
\[
\dfrac{\partial w_{(i,k)}}{\partial a_{(i,j)}} = w_{(i,k)} \left( \delta_{k,j} - w_{(i,j)} \right),
\]
where \(\delta_{k,j}\) is the Kronecker delta, which is 1 if \(k = j\) and 0 otherwise.

2. Compute \(\dfrac{\partial \theta_i}{\partial a_{(i,j)}}\):

Since \(\theta_i\) depends on \(w_{(i,j)}\):
\[
\dfrac{\partial \theta_i}{\partial a_{(i,j)}} = \sum_{k=1}^d \dfrac{\partial \theta_i}{\partial w_{(i,k)}} \cdot \dfrac{\partial w_{(i,k)}}{\partial a_{(i,j)}} = \sum_{k=1}^d \Theta_k \cdot \dfrac{\partial w_{(i,k)}}{\partial a_{(i,j)}}.
\]

3. Use the chain rule to compute \(\dfrac{\partial L}{\partial a_{(i,j)}}\):
\[
\dfrac{\partial L}{\partial a_{(i,j)}} = \langle\dfrac{\partial \ell}{\partial \theta_i} , \dfrac{\partial \theta_i}{\partial a_{(i,j)}} \rangle= \langle\dfrac{\partial \ell}{\partial \theta_i} , \sum_{k=1}^d \Theta_k \cdot \dfrac{\partial w_{(i,k)}}{\partial a_{(i,j)}}\rangle.
\]

4. Substitute \(\dfrac{\partial w_{(i,k)}}{\partial a_{(i,j)}}\):
\[
\dfrac{\partial L}{\partial a_{(i,j)}} = \langle\dfrac{\partial \ell}{\partial \theta_i} ,\sum_{k=1}^d \Theta_k \cdot w_{(i,k)} \left( \delta_{k,j} - w_{(i,j)} \right)\rangle.
\]

Simplify the expression:
\begin{align*}
\dfrac{\partial L}{\partial a_{(i,j)}} &= \langle\dfrac{\partial \ell}{\partial \theta_i} ,\left[ \Theta_j w_{(i,j)} (1 - w_{(i,j)}) - \sum_{k \neq j} \Theta_k w_{(i,k)} w_{(i,j)} \right] \rangle\\
&= \langle\dfrac{\partial \ell}{\partial \theta_i}, w_{(i,j)} \left( \Theta_j - \sum_{k=1}^d \Theta_k w_{(i,k)} \right)\rangle.
\end{align*}
Note that \(\sum_{k=1}^d \Theta_k w_{(i,k)} = \theta_i\), so:
\[
\dfrac{\partial L}{\partial a_{(i,j)}} = \langle\dfrac{\partial \ell}{\partial \theta_i} ,w_{(i,j)} \left( \Theta_j - \theta_i \right)\rangle.
\]

\subsection{Final Expressions}

Gradient with respect to \(\Theta_j\):
\[
\dfrac{\partial L}{\partial \Theta_j} = \sum_{i=1}^m w_{(i,j)} \cdot \dfrac{\partial \ell}{\partial \theta_i}.
\]

Gradient with respect to \(a_{(i,j)}\):
\[
\dfrac{\partial L}{\partial a_{(i,j)}} = w_{(i,j)}\cdot\langle\dfrac{\partial \ell}{\partial \theta_i} , \left( \Theta_j - \theta_i \right)\rangle.
\]

These expressions efficiently compute the gradients while respecting the constraint on \(w_{(i,j)}\). The term \(\Theta_j - \theta_i\) represents the difference between the parameter \(\Theta_j\) and the weighted average \(\theta_i\), scaled by \(w_{(i,j)}\) and the derivative of the loss with respect to \(\theta_i\). We replace Eq~\eqref{eq:gradient_theta} and Eq~\eqref{eq:gradient_w} with the above to equations to perform a row softmax constraint on merging weights.


\clearpage 
\onecolumn
\section{Convergence Analysis}
\label{sec:convergence_analysis}

\subsection{Modelling Assumptions}

\begin{description}[leftmargin=2.2em,labelsep=0.6em]
\item[\textbf{A1. $L$–smooth objective}]%
      The global loss
      $F(\Theta,w)$ is $L$–smooth w.r.t.\ both $\Theta$ and $w$:
      \[
        \|\nabla F(u)-\nabla F(\bar u)\|
        \;\le\;L\|u-\bar u\|,
        \quad
        u=(\Theta,w).
      \]

\item[\textbf{A2. Bounded local gradients}]%
      For every client~$i$ and any model parameter,
      $\|\nabla F_i(\theta)\|\le G$.

\item[\textbf{A3. Row–simplex router weights}]%
      Each client weight vector lies on the probability simplex  
      $\displaystyle
        \sum_{j=1}^{d}w_{i,j}^t = 1,\;
        w_{i,j}^t\in[0,1]$.

\item[\textbf{A4. Bounded expert norms}]%
      $\|\Theta_j^{t}\| \le C_\Theta$ for every expert~$j$ and round~$t$.

\item[\textbf{A5. Heterogeneity bound}]%
      Global vs.\ local gradient gap is uniformly bounded:
      $\|\nabla F(\theta)-\nabla F_i(\theta)\|\le\zeta$.

\end{description}

\vspace{.5em}
\subsection{Key Lemmas and Theorems}

\begin{description}[leftmargin=2.2em,labelsep=0.6em]
\item[\textbf{L1 (Server descent)}]  
      \emph{[Eq.\,\eqref{eq:final-descent}]}  
      Updating $\Theta,w$ by full gradients with step $\eta<2/L$ yields
      \[
        F(\Theta^{t+1},w^{t+1})
        \;\le\;
        F(\Theta^{t},w^{t})
        -\eta\!\Bigl(1-\tfrac{\eta L}{2}\Bigr)
        \bigl(m+dC_\Theta^{2}\bigr)\!
        \sum_{i=1}^{m}\|g_i^{t}\|^{2}.
      \]

\item[\textbf{L2 (Client deviation)}]  
      \emph{[Eq.\,\eqref{eq:client_final}]}  
      Let $g_i^{t}$ be the idea gradient on client $i$ round $t$,  $\bar g_i^{t}$ be the true gradient. The uploaded gradient on client~$i$ satisfies
      \[
        g_i^{t}
        \;=\;
       \bar g_i^{t}
        \;+\;
        \delta
        ,\qquad
        \delta=\zeta+\eta_{\mathrm{loc}}LGE/2.
      \]



\item[\textbf{T1 (Convergence Rate)}]  
       \emph{[Eq.\,\eqref{eq:Finite-Time Rate}]},
      \[
        \frac1T\sum_{t<T}\sum_{i}\|\bar g_i^{t}\|^{2}
        =\mathcal O\!\bigl(1/(Td)\bigr).
      \]
      The average squared gradient norm decreases at a convergence rate that is inversely proportional to both the number of training rounds $T$ and the number of global models $d$.

\item[\textbf{T2 (Asymptotic neighbourhood)}]  
      \emph{[Eq.\,\eqref{eq:neighborhood}]}  
      In general,
      \[
\limsup_{T \to \infty}
\frac{1}{T} \sum_{t=0}^{T-1} \sum_{i=1}^{m}
\left\| \frac{1}{E} \sum_{e=0}^{E-1} \nabla F_i\left( \theta_i^{t,e} \right) \right\|^2
= \mathcal{O} \left(
    m \left( \zeta + \eta_{\mathrm{loc}} L G E \right)^2
\right)
      \]
      The algorithm converges to a stationary‑point neighbourhood whose
      radius grows with data heterogeneity~$\zeta$ and local‑update drift
      $(\eta_{\mathrm{loc}}E)^2$.
\end{description}

\smallskip
\noindent
\textbf{Interpretation.}\;  
The average squared gradient norm decreases at a convergence rate that is inversely proportional to both the number of training rounds $T$ and the number of global models $d$—that is, it shrinks proportionally to 
$O\!\bigl(1/(Td)\bigr)$.The method stabilises inside a ball whose size is quadratic in
\(\zeta\) and \((\eta_{\mathrm{loc}}E)^2\).

\subsection{Bounding the Objective Descent in Server Update}

We consider the optimization objective \( F(\Theta, w)=\min_{\Theta,w} \sum_{i=1}^m \frac{n_i}{n} \ell \left(Y_i, f(X_i; \sum_{j=1}^d w_{(i,j)} \cdot \Theta_j)\right) \)with smoothness constant \( L \), and we analyze the descent of the objective under gradient updates. Let \( \Delta\Theta = \Theta^{t+1} - \Theta^t \), \( \Delta w = w^{t+1} - w^t \), and suppose \( F \) is \( L \)-smooth in both \( \Theta \) and \( w \). Then, the standard second-order upper bound yields:
\begin{align}
F(\Theta^{t+1}, w^{t+1}) &\le F(\Theta^{t}, w^{t}) 
+ \left\langle \nabla_{\Theta} F, \Delta\Theta \right\rangle 
+ \left\langle \nabla_{w} F, \Delta w \right\rangle 
+ \frac{L}{2} \left( \|\Delta\Theta\|^{2} + \|\Delta w\|^{2} \right). \label{eq:smooth-bound}
\end{align}

Assume both \( \Theta \) and \( w \) are updated via gradient descent with learning rate \( \eta \):
\[
\Delta\Theta = -\eta \nabla_{\Theta} F(\Theta^{t}, w^{t}), \quad 
\Delta w = -\eta \nabla_{w} F(\Theta^{t}, w^{t}).
\]

Substitute into \eqref{eq:smooth-bound}, we obtain:
\begin{align}
F(\Theta^{t+1}, w^{t+1}) 
&\le F(\Theta^{t}, w^{t}) 
- \eta \left\| \nabla_{\Theta} F(\Theta^{t}, w^{t}) \right\|^{2}
- \eta \left\| \nabla_{w} F(\Theta^{t}, w^{t}) \right\|^{2} \notag\\
&\quad + \frac{L \eta^{2}}{2} 
\left( \left\| \nabla_{\Theta} F(\Theta^{t}, w^{t}) \right\|^{2}
+ \left\| \nabla_{w} F(\Theta^{t}, w^{t}) \right\|^{2} \right) \notag\\
&= F(\Theta^{t}, w^{t}) 
- \eta\left(1 - \frac{\eta L}{2} \right)
\left( \left\| \nabla_{\Theta} F(\Theta^{t}, w^{t}) \right\|^{2}
+ \left\| \nabla_{w} F(\Theta^{t}, w^{t}) \right\|^{2} \right). \label{eq:descent}
\end{align}

Let us define the quantity:
\[
\star := 
\left\| \nabla_{\Theta} F(\Theta^{t}, w^{t}) \right\|^{2}
+ \left\| \nabla_{w} F(\Theta^{t}, w^{t}) \right\|^{2}.
\]

We now compute this term based on the gradient structure. As in Eq.~\ref{eq:gradient_theta},Eq.~\ref{eq:gradient_w}, the partial gradients are:
\begin{align}
\nabla_{\Theta_j} F &= \sum_{i=1}^{m} w_{i,j}^{t} \, g_i^{t}, \label{eq:grad-theta} \\
\nabla_{w_{i,j}} F &= \left\langle \Theta_j^{t}, g_i^{t} \right\rangle. \label{eq:grad-w}
\end{align}

Then, the quantity \( \star \) becomes:
\begin{align}
\star 
&= \sum_{j=1}^{d} 
\left\| \sum_{i=1}^{m} w_{i,j}^{t} \, g_i^{t} \right\|^{2}
+ \sum_{i=1}^{m} \sum_{j=1}^{d} 
\left\langle \Theta_j^{t}, g_i^{t} \right\rangle^{2} \notag\\
&=: a + b. \label{eq:star-split}
\end{align}

We estimate each term separately.

\subsection*{Bounding \( a \)}

We apply the Cauchy–Schwarz inequality to each vector-valued sum:
\begin{align}
a 
&= \sum_{j=1}^{d} 
\left\| \sum_{i=1}^{m} \sqrt{w_{i,j}^{t}} \cdot \left(\sqrt{w_{i,j}^{t}}\, g_i^{t} \right) \right\|^{2} \notag\\
&\le \sum_{j=1}^{d} 
\left( \sum_{i=1}^{m} w_{i,j}^{t} \right) 
\left( \sum_{i=1}^{m} w_{i,j}^{t} \|g_i^{t}\|^{2} \right) \label{eq:cauchy-a}
\end{align}

Assume each row of \( w \) satisfies the simplex constraint:
\[
\sum_{j=1}^{d} w_{i,j}^{t} = 1, \quad 0 \le w_{i,j}^{t} \le 1.
\]

Then we can upper-bound:
\begin{align}
a 
&\le m \sum_{j=1}^{d} \sum_{i=1}^{m} w_{i,j}^{t} \|g_i^{t}\|^{2} \notag\\
&= m \sum_{i=1}^{m} \left( \sum_{j=1}^{d} w_{i,j}^{t} \right) \|g_i^{t}\|^{2} \notag\\
&= m \sum_{i=1}^{m} \|g_i^{t}\|^{2} \label{eq:bound-a}
\end{align}

\subsection*{Bounding \( b \)}

Assume \( \|\Theta_j^{t}\| \le C_{\Theta} \) for all \( j \). Then:
\begin{align}
b 
&= \sum_{i=1}^{m} \sum_{j=1}^{d} 
\left\langle \Theta_j^{t}, g_i^{t} \right\rangle^{2} \notag\\
&\le \sum_{i=1}^{m} \sum_{j=1}^{d} 
\|\Theta_j^{t}\|^{2} \cdot \|g_i^{t}\|^{2} \notag\\
&\le d C_{\Theta}^{2} \sum_{i=1}^{m} \|g_i^{t}\|^{2}. \label{eq:bound-b}
\end{align}

\subsection*{Combining \( a \) and \( b \)}

Summing \eqref{eq:bound-a} and \eqref{eq:bound-b}, we obtain:
\begin{align}
\star = a + b 
\le \left(m + dC_{\Theta}^{2} \right) \sum_{i=1}^{m} \|g_i^{t}\|^{2}. \label{eq:final-star}
\end{align}

\subsection*{Final Descent Inequality}

Substitute \eqref{eq:final-star} into \eqref{eq:descent}, we conclude:

\begin{equation}
\boxed{
F(\Theta^{t+1}, w^{t+1}) 
\le F(\Theta^{t}, w^{t}) 
- \eta \left(1 - \frac{\eta L}{2} \right) 
\left(m + dC_{\Theta}^{2} \right) 
\sum_{i=1}^{m} \|g_i^{t}\|^{2}.
}\label{eq:final-descent}
\end{equation}

\subsection{Deviation Bound Between Actual and Ideal Gradients on Clients}

We analyze the deviation between the actual accumulated local gradient on client $i$. The actual accumulated average local gradient over $E$ local steps is denoted by
\[
\bar{g}_i^{t} := \frac{1}{E} \sum_{e=0}^{E-1} \nabla F_i\left( \theta_i^{t,e} \right),
\]
and the ideal gradient \( g_i^{t} := \nabla F(\theta_i^{t,0}) \) evaluated at the initial local model before local updates. The goal is to upper bound the difference:
\[
\left\| g_i^{t} - \bar{g}_i^{t} \right\|.
\]

By adding and subtracting appropriate terms, we decompose:
\begin{align}
g_i^{t}
&= \nabla F(\theta_i^{t,0}) \notag\\
&= \frac{1}{E} \sum_{e=0}^{E-1} \nabla F_i(\theta_i^{t,e})
+ \left[ \nabla F(\theta_i^{t,0}) - \nabla F_i(\theta_i^{t,0}) \right]
+ \left[ \nabla F_i(\theta_i^{t,0}) - \frac{1}{E} \sum_{e=0}^{E-1} \nabla F_i(\theta_i^{t,e}) \right] \label{eq:grad-decomp}
\end{align}

Taking norm and applying triangle inequality:
\begin{align}
\left\| g_i^{t} - \bar{g}_i^{t} \right\|
&\le \left\| \nabla F(\theta_i^{t,0}) - \nabla F_i(\theta_i^{t,0}) \right\|
+ \left\| \nabla F_i(\theta_i^{t,0}) - \frac{1}{E} \sum_{e=0}^{E-1} \nabla F_i(\theta_i^{t,e}) \right\| \label{eq:tri-split}
\end{align}

Assume bounded heterogeneity between global and local gradients:
\begin{equation}
\left\| \nabla F(\theta) - \nabla F_i(\theta) \right\| \le \zeta, \quad \forall \theta.
\end{equation}

Then the first term in \eqref{eq:tri-split} is bounded by \( \zeta \). For the second term, use $L$-smoothness of \( F_i \):
\begin{align}
\left\| \nabla F_i(\theta_i^{t,0}) - \frac{1}{E} \sum_{e=0}^{E-1} \nabla F_i(\theta_i^{t,e}) \right\|
&\le \frac{1}{E} \sum_{e=0}^{E-1} 
\left\| \nabla F_i(\theta_i^{t,0}) - \nabla F_i(\theta_i^{t,e}) \right\| \notag\\
&\le \frac{1}{E} \sum_{e=0}^{E-1} L \left\| \theta_i^{t,0} - \theta_i^{t,e} \right\| \label{eq:smooth-local}
\end{align}

Assuming local update with learning rate \( \eta_{\mathrm{loc}} \), we write:
\begin{equation}
\theta_i^{t,e} = \theta_i^{t,0} - \eta_{\mathrm{loc}} \sum_{s=0}^{e-1} \nabla F_i(\theta_i^{t,s})
\end{equation}

Assume each local gradient is bounded by \( \|\nabla F_i(\theta)\| \le G \), then:
\begin{align}
\left\| \theta_i^{t,0} - \theta_i^{t,e} \right\|
= \left\| \eta_{\mathrm{loc}} \sum_{s=0}^{e-1} \nabla F_i(\theta_i^{t,s}) \right\|
\le \eta_{\mathrm{loc}} \sum_{s=0}^{e-1} \left\| \nabla F_i(\theta_i^{t,s}) \right\|
\le \eta_{\mathrm{loc}} G e
\end{align}

Plug into \eqref{eq:smooth-local}:
\begin{align}
\left\| \nabla F_i(\theta_i^{t,0}) - \frac{1}{E} \sum_{e=0}^{E-1} \nabla F_i(\theta_i^{t,e}) \right\|
&\le \frac{1}{E} \sum_{e=0}^{E-1} L \eta_{\mathrm{loc}} G e 
= \eta_{\mathrm{loc}} L G \cdot \frac{1}{E} \sum_{e=0}^{E-1} e 
= \eta_{\mathrm{loc}} L G \cdot \frac{E-1}{2}
\end{align}

Combining both parts of \eqref{eq:tri-split}, we obtain the deviation bound:
\begin{equation}
\left\| g_i^{t} - \bar{g}_i^{t} \right\|
\le \zeta + \eta_{\mathrm{loc}} L G \cdot \frac{E-1}{2}
\end{equation}

\begin{equation}\
\boxed{
g_i^{t}
= \bar{g}_i^{t} + \zeta + \eta_{\mathrm{loc}} L G \cdot \tfrac{E-1}{2}
}\label{eq:client_final}
\end{equation}

\subsection{Convergence of FedMerge}

We derive an upper bound on the average squared gradient norm over $T$ rounds by combining server-side descent with client-side gradient deviation.

\subsection*{One-step Descent with Client Approximation}

Let 
\[
\delta := \zeta + \eta_{\mathrm{loc}} L G \cdot \frac{E - 1}{2},
\]
which accounts for gradient deviation due to heterogeneity and local drift. Substituting 
\[
g_i^t = \bar{g}_i^t + \delta
\]
into the descent inequality yields:
\begin{align}
F(\Theta^{t+1}, w^{t+1})
&\le F(\Theta^{t}, w^{t})
- \eta\left(1 - \frac{\eta L}{2}\right)
d (m + C_\Theta^2)
\sum_{i=1}^{m} \left\| \bar{g}_i^t + \delta \right\|^2. \label{eq:onestep-descent}
\end{align}

\subsection*{Lower Bound via Cauchy–Schwarz}

\begin{align}
\left\| \bar{g}_i^t + \delta \right\|^2
&= \left\| \bar{g}_i^t \right\|^2 
+ 2 \left\langle \bar{g}_i^t, \delta \right\rangle 
+ \left\| \delta \right\|^2 \notag\\
&\ge \left\| \bar{g}_i^t \right\|^2 
- 2 \left\| \bar{g}_i^t \right\| \left\| \delta \right\|
+ \left\| \delta \right\|^2 \notag\\
&\ge \left\| \bar{g}_i^t \right\|^2 - (2G-\delta)\delta. \label{eq:cauchy-lower}
\end{align}

\subsection*{Substituting the Bound}

Inserting \eqref{eq:cauchy-lower} into \eqref{eq:onestep-descent}, we have:
\begin{align}
F(\Theta^{t+1}, w^{t+1})
&\le F(\Theta^{t}, w^{t})
- \eta\left(1 - \frac{\eta L}{2} \right)
d (m + C_\Theta^2)
\sum_{i=1}^{m} \left\| \bar{g}_i^t \right\|^2 \notag\\
&\quad + \eta\left(1 - \frac{\eta L}{2} \right)
d (m + C_\Theta^2) m (2G-\delta)\delta. \label{eq:descent-final}
\end{align}

\subsection*{Telescoping Over Rounds}

Summing \eqref{eq:descent-final} over $t = 0$ to $T - 1$:
\begin{align}
F(\Theta^T, w^T)
&\le F(\Theta^0, w^0)
- \eta\left(1 - \frac{\eta L}{2} \right)
d (m + C_\Theta^2)
\sum_{t=0}^{T-1} \sum_{i=1}^{m} \left\| \bar{g}_i^t \right\|^2 \notag\\
&\quad + \eta\left(1 - \frac{\eta L}{2} \right)
d (m + C_\Theta^2) m T (2G-\delta)\delta. \label{eq:telescope}
\end{align}

Divide \eqref{eq:telescope} by \( T \cdot \eta\left(1 - \frac{\eta L}{2} \right)d (m + C_\Theta^2) \), we obtain:
\begin{align}
\frac{1}{T}\sum_{t=0}^{T-1}\sum_{i=1}^{m}\|\bar g_i^{t}\|^{2}
&\le
\frac{F(\Theta^0,w^0)-F(\Theta^{T},w^{T})}{\eta\left(1-\frac{\eta L}{2}\right)
      d(m+C_\Theta^{2})\,T}
+ m(2G-\delta)\delta \\[4pt]
&\le
\frac{\Delta F}{\eta\left(1-\frac{\eta L}{2}\right)(m+C_\Theta^{2})}\cdot\frac{1}{Td}
+ m(2G-\delta)\delta. \label{eq:avg-grad-final}
\end{align}

\eqref{eq:avg-grad-final} holds When $\eta<\frac{2}{L}$ and $\eta_{\mathrm{loc}}<\frac{2G-\zeta}{LG\frac{E-1}{2}}$.

\vspace{1ex}
\paragraph{Corollary 1 (Finite-Time Rate).}
When $T$ is finite, the algorithm enjoys the ergodic rate:
\begin{equation}
\boxed{
\frac{1}{T}\sum_{t=0}^{T-1}\sum_{i=1}^{m}\|\bar g_i^{t}\|^{2}
= \mathcal{O}\left(\tfrac{1}{T d}\right)
}\label{eq:Finite-Time Rate}
\end{equation}

\vspace{1ex}
\paragraph{Corollary 2 (Asymptotic Neighborhood).}
As $T \to \infty$, the first term vanishes, and we obtain:
\begin{equation}
\limsup_{T \to \infty} 
\frac{1}{T} \sum_{t=0}^{T-1} \sum_{i=1}^{m} \|\bar g_i^t\|^2
\le m(2G-\delta)\delta. \label{eq:limsup-bound}
\end{equation}

Substituting 
\[
\delta = \zeta + \eta_{\mathrm{loc}} L G \cdot \frac{E - 1}{2},
\]
we have:
\[
\limsup_{T \to \infty} 
\frac{1}{T} \sum_{t=0}^{T-1} \sum_{i=1}^{m}
\left\| \bar{g}_i^t \right\|^2
\le m\left(2G-\zeta - \eta_{\mathrm{loc}} L G \cdot \frac{E - 1}{2}\right)\left( \zeta + \eta_{\mathrm{loc}} L G \cdot \frac{E - 1}{2} \right).
\]

Equivalently, in terms of the average local gradients:
\begin{equation}
\boxed{
\limsup_{T \to \infty}
\frac{1}{T} \sum_{t=0}^{T-1} \sum_{i=1}^{m}
\left\| \frac{1}{E} \sum_{e=0}^{E-1} \nabla F_i\left( \theta_i^{t,e} \right) \right\|^2
= \mathcal{O} \left(
    m \left( \zeta + \eta_{\mathrm{loc}} L G E \right)^2
\right)
} \label{eq:neighborhood}
\end{equation}

This shows FeMAM converges to a neighborhood around stationary points, whose size is determined by $\zeta$ and $E$.

\end{document}